\documentclass[10pt,twocolumn,letterpaper]{article}

\usepackage{cvpr}              % To produce the CAMERA-READY version

\usepackage{multirow}
\usepackage{lscape}
\usepackage{adjustbox}
\usepackage{booktabs}
\usepackage{array}
\usepackage{arydshln}
\usepackage{tabularx}
\usepackage{rotating}
\usepackage{diagbox}
\usepackage{graphicx}
\usepackage{caption}
\usepackage{subcaption}

\DeclareUnicodeCharacter{0421}{C}

\definecolor{cvprblue}{rgb}{0.21,0.49,0.74}
\usepackage[pagebackref,breaklinks,colorlinks,allcolors=cvprblue]{hyperref}

\def\paperID{14} % *** Enter the Paper ID here
\def\confName{CVPR}
\def\confYear{2025}

\title{BeetleVerse: A study on taxonomic classification of ground beetles}

\author{
    \textbf{S~M~Rayeed}\thanks{Corresponding author: rayees@rpi.edu}\\
    Rensselaer Polytechnic Institute\\
    \and
    Alyson~East\\
    The University of Maine\\
    \and
    Samuel~Stevens\\
    The Ohio State University\\
    \and
    Sydne~Record\\
    The University of Maine\\
    \and
    Charles~V.~Stewart\\
    Rensselaer Polytechnic Institute\\
}

\makeatletter
\def\ps@cvprfooter{%
  \def\@oddfoot{
    \parbox{\textwidth}{
      {\small\textcolor{gray}{
      42nd Conference on Computer Vision and Pattern Recognition (CVPR 2025)\\
      Workshop: 5th Workshop on CV4Animals: Computer Vision for Animal Behavior Tracking and Modeling
      }}
    }
  }
}
\makeatother
\begin{document}

\maketitle
\thispagestyle{cvprfooter}

% PREVIOUS ABSTRACT
% \begin{abstract}
% Ground beetles play a vital role in monitoring biodiversity and ecosystem health, but their taxonomic identification is hindered by subtle morphological differences and insufficient data. Manual expert-driven approaches struggle to keep up with the demands of large-scale beetle studies, leading to significant bottlenecks. To overcome these challenges, in this paper, we introduce Carabids100k, a of vast taxonomic diversity, spanning over 230 genera and 1769 species. We evaluate 12 vision-language models across multiple key dimensions, including zero-shot prediction, genus and species-level identification, data optimization, cross-domain adaptation, and multimodal feature integration. Our benchmarking results indicate the Vision and Language Transformer as the best model, with 97.01\% accuracy at genus and 93.94\% at species level \textcolor{red}{classification}; while other models also significantly outperform baseline predictions. We also optimize efficiency by reducing data \underline{overhead} up to 50\%, and improve accuracy through multimodal integration of visual, morphological, and environmental data. By addressing relatively poor performance for in-situ data, we highlight a critical domain gap in adapting to real-world imagery. In short, our findings provide a foundation for future advancements in automated large-scale taxonomic identification, and identify critical gaps for future work. 
% \end{abstract}

\begin{abstract}
Ground beetles are a highly sensitive and speciose biological indicator, making them vital for monitoring biodiversity. However, they are currently an underutilized resource due to the manual effort required by taxonomic experts to perform challenging species differentiations based on subtle morphological differences, precluding widespread applications. In this paper, we evaluate 12 vision models on taxonomic classification across four diverse, long-tailed datasets spanning over 230 genera and 1769 species, with images ranging from controlled laboratory settings to challenging field-collected (in-situ) photographs. We further explore taxonomic classification in two important real-world contexts: sample efficiency and domain adaptation. Our results show that the Vision and Language Transformer combined with an MLP head is the best performing model, with 97\% accuracy at genus and 94\% at species level. Sample efficiency analysis shows that we can reduce train data requirements by up to 50\% with minimal compromise in performance. The domain adaptation experiments reveal significant challenges when transferring models from lab to in-situ images, highlighting a critical domain gap. Overall, our study lays a foundation for large-scale automated taxonomic classification of beetles, and beyond that, advances sample-efficient learning and cross-domain adaptation for diverse long-tailed ecological datasets.
\end{abstract}

\section{Introduction}
\label{sec:introduction}
Ground beetles (family \textit{Carabidae}; commonly known as carabids) represent one of the largest and most diverse families of beetles, comprising over 40,000 described species worldwide \cite{lovei1996ecology}. This diverse lineage serves as pivotal bio-indicators of environmental health and natural pest regulators in agricultural ecosystems, highlighting the importance of their accurate taxonomic classification for tracking biodiversity changes, monitoring invasive species, and evaluating ecosystem resilience \cite{rainio2003ground}. Accurate classification remains a major challenge, and therefore an interesting computer vision problem, due to several factors: (1) subtle morphological distinctions between closely related species, often requiring sub-millimeter scrutiny of elytral striations, pronotum curvature, or antennal segmentation \cite{ribera1999comparative}; (2) substantial intraspecies variation across geographic clines, life stages, and environmental conditions \cite{sukhodolskaya2016intra}; (3) an overwhelming  taxonomic diversity \cite{blair2020robust}; and (4) lack of well-curated datasets to train computer vision models for automated classification \cite{koivula2011useful}.
\begin{figure}[!t]
    \centering
    \fbox{\includegraphics[width=0.22\columnwidth,height=4cm,keepaspectratio]{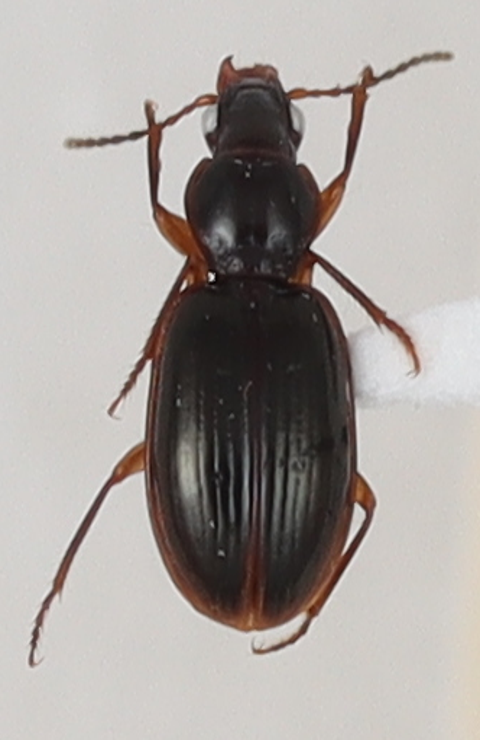}}\hfill
    \fbox{\includegraphics[width=0.22\columnwidth,height=4cm,keepaspectratio]{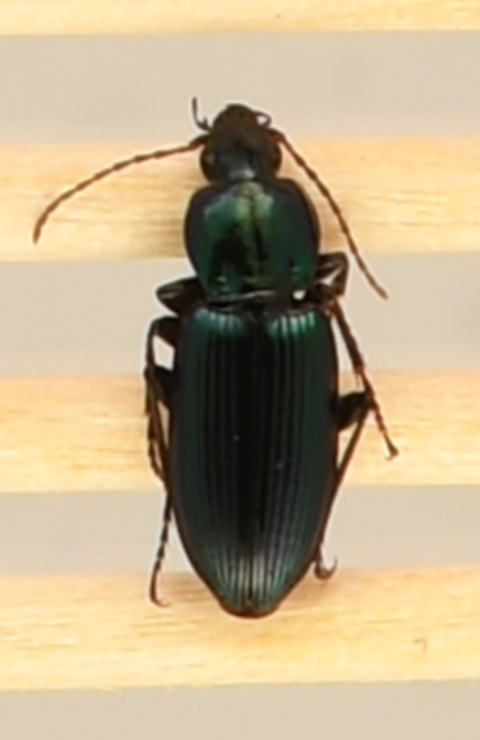}}\hfill
    \fbox{\includegraphics[width=0.22\columnwidth,height=4cm,keepaspectratio]{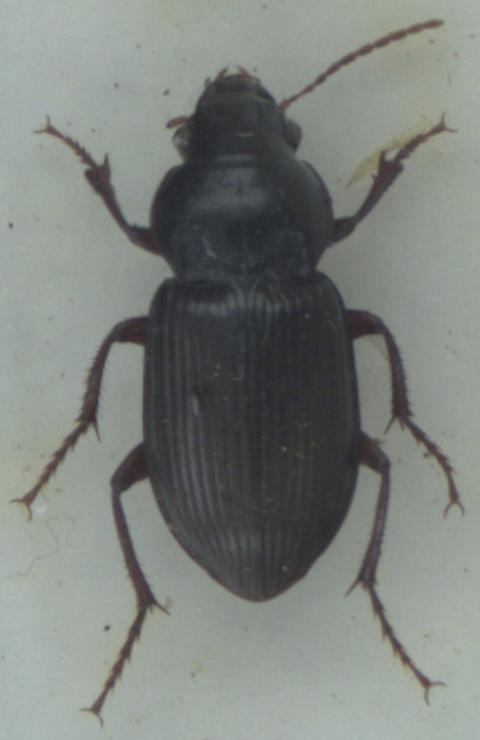}}\hfill
    \fbox{\includegraphics[width=0.22\columnwidth,height=4cm,keepaspectratio]{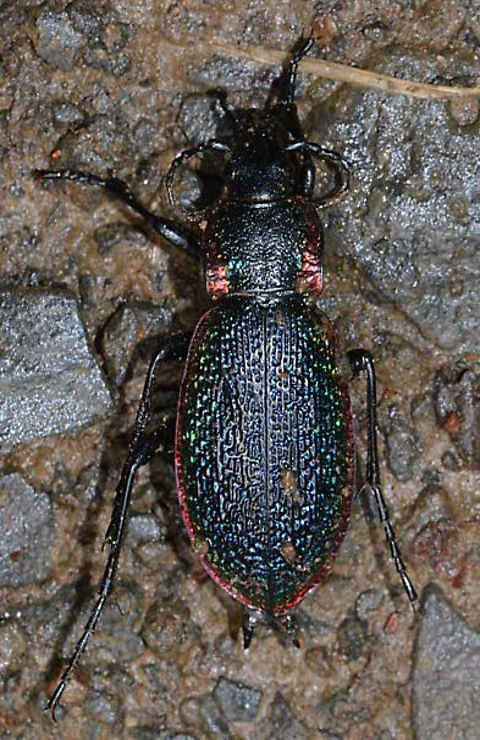}}
    \captionsetup{skip=3pt}
    \caption{Samples from the four datasets described in Section~\ref{subsec:DataCollection}. From left: \textit{Mecyclothroax konanus} (from BeetlePUUM), \textit{Poecilus scitulus} (from BeetlePalooza), \textit{Amara aulica} (from NHM-Carabids), and \textit{Carabus vietinghoffii} (from I1MC).}
    \label{fig:beetles}
\end{figure}
Existing workflows depend on taxonomists using their domain-specific knowledge to analyze subtle appendage characteristics that often require both dorsal and ventral views of a specimen \cite{hoekman2017design, evans2023lives, evans2000inordinate, stovces2025multilevel}. This manual and labor-intensive  process severely constrains large-scale taxonomic studies. To address these gaps, we
\begin{enumerate}
    \item Conduct a comprehensive study of 12 state-of-the-art vision-only and vision-language models, evaluating their effectiveness in hierarchical taxonomic classification through linear probing, comparing performance to each other and to a zero-shot baselines.
    \item Investigate sample efficiency, a crucial factor for long-tailed datasets, by analyzing how accuracy varies with dataset size and assessing the balance between training volume and performance. Our findings reveal a trade-off where strategic subsampling can achieve strong performance and reduce costs - a major consideration for studies with long-tailed data and limited resources.
    \item Assess cross-dataset domain adaptation, investigating whether pretrained vision models generalize well across datasets, particularly from curated collection images to in-situ field images. This challenge remains fundamental to ecology given the higher classification accuracy and greater volume of data in lab collections as compared to in-situ data, and the potential benefits of reducing destructive sampling in favor of collecting in-situ images.
        
\end{enumerate}
Benchmarking results show that Vision and Language Transformer (ViLT) \cite{kim2021vilt} achieves the highest accuracy of 97.15\% at genus level and 93.97\% at species, substantially outperforming zero-shot baselines. For sample efficiency, classification accuracy remains largely unchanged with 50\% data.
% , while maintaining sub-optimal accuracy CHUCK: WHAT DOES SUBOPTIMAL MEAN HERE???  with just 20\% data.
% , effectively reducing computational overhead by 80\%.
Domain shift scores indicate that while models generalize well across controlled lab settings, a performance drop of 37\% for in-situ images highlights the need for robust domain adaptation.
% \textcolor{blue}{Suggestion: maybe keep this one sentence? Additionally, we conduct some experiments on multimodal data integration, where we analyze the impact of augmenting morphological traits and environmental metadata alongside visual features for enhanced  classification, described in Appendix XX.}
% Lastly, multimodal integration results show that vision-only models struggle with added modalities when data is limited but benefit at scale, whereas vision-language models perform better.
In short, our work introduces a novel approach to streamline taxonomic classification of carabids, while identifying critical gaps in domain shift.

\section{Related Work}
\label{sec:related_work}

The computational identification of insects has emerged as a critical research domain, with numerous studies addressing the challenges of taxonomic classification through advanced machine learning (ML) techniques \cite{nanni2025insect, weeks1999automating, badirli2023classifying, xia2015automatic, tannous2023deep, bjerge2023accurate, kasinathan2021insect}. Domain adaptation is critical \cite{yang2020phase, shi2024potential, kim2024application, fujisawa2023image, majewski2023mixing}. For example, \cite{shi2024potential} investigated the generalization of ML models across different imaging conditions, revealing significant performance drops for out of distribution data. Several works proposed multimodal approaches integrating visual features with ecological metadata \cite{cumming2002evolution, higham2013introduction, ariznavarreta2024multimodal, Rayeed_2025_ICCV}, showing different signal modalities often evolve independently to serve distinct functions.
A few recent works have explored automated methods for identifying and monitoring carabids \cite{blair2020robust, hoekman2017design, hansen2020species, liu2025beetleflow, rayeed2026continental, wu2019deep, abeywardhana2021deep}. Most notable among them is NEON's continental-scale ground beetle monitoring program \citep{hoekman2017design, NEON-pinned-specimens, NEON-pinned-beetles-metadata, Portal2022-ho, Portal2022-qu}. Among others, \cite{blair2020robust} achieved 84.6\% species-level accuracy using linear discriminant analysis on pitfall trap-collected carabids; \cite{hansen2020species} achieved 51.9\% species-level accuracy using applied neural networks; and \cite{abeywardhana2021deep} got 90\% accuracy on Sri Lankan beetles by using background removal, segmentation, and transfer learning. However, these approaches are limited by small taxonomic coverage and the use of conventional ML models that lack the representational power of cutting-edge vision encoders. Our work addresses these limitations by leveraging that power and covering a broader taxonomic range.

\section{Methodology}
\label{sec:methodology}
% This section covers a brief summary and statistical analysis of the datasets used, followed by the methods used for benchmarking, sample efficient probing, and domain adaptation.

\subsection{Datasets used in this study}
\label{subsec:DataCollection}
We use four distinctive datasets, comprised of 100,885 carabid images in total, spanning over 230 genera and 1769 species. The datasets are as follows: 
\begin{itemize}
    \item BeetlePUUM: A dataset of Hawaiian-endemic carabids, imaged from pinned specimens at the Pu'u Maka'ala site (PUUM) of National Ecological Observatory Network (NEON) \cite{rayeed2025HawaiiBeetles}. It contains high quality images taken in controlled settings and is rich with ecological metadata, though limited in taxonomic diversity.
    \item BeetlePalooza: Another dataset of NEON carabids, from 30 sites across the continental US, containing images of beetles preserved in ethanol vials \cite{Fluck2018_NEON_Beetle}. Although it broadens taxonomic and geographic diversity (36 genera, 76 species) and contains ecological metadata, the specimens vary in spacing and orientation due to the collection's scale and the fragile nature of the specimens.
    \item NHM-Carabids: A collection from the Natural History Museum in London \cite{hansen_2019_3549369}, with expert-verified taxonomy labels, no accompanying metadata, and containing some images that are blurry or show lighting inconsistencies.
    \item I1MC: A filtered subset of carabids from the Insect-1M dataset \cite{nguyen2024insect}. It includes both lab specimens and in-situ images, and offers taxonomic diversity but lacks consistency, with variable image quality, inconsistent specimen orientation, and frequent partial captures instead of complete specimens.      
\end{itemize}
Apart from experimenting on these datasets individually, we also put together a merged version combining all four to facilitate large-scale analyses across diverse imaging conditions, taxonomic groups, and ecological contexts. Table~\ref{tab:dataset_statistics} provides detailed statistics for each dataset and the merged collection. For more details, see Appendix ~\ref{app_subsec:data-details}.
\begin{table}[h]
\centering
\setlength{\tabcolsep}{5pt}
\resizebox{\columnwidth}{!}{%
\begin{tabular}{l*{6}{c}}
\toprule
\textbf{Data} & \textbf{No. of} & \textbf{No. of} & \textbf{No. of} & \textbf{Genus} & \textbf{Species} & \textbf{T+E}\\
\textbf{Code} & \textbf{Images} & \textbf{Genera} & \textbf{Species} & \textbf{N/A} & \textbf{N/A} & \textbf{Data} \\
\midrule
\midrule
BPM & 1803 & 4 & 14 & 0 & 0 & Yes \\ 
BPZ & 11399 & 36 & 76 & 17 & 27 & Yes \\ 
NHMC & 63077 & 77 & 290 & 0 & 0 & No \\ 
I1MC & 24606 & 206 & 1531 & 424 & 4328 & No \\
\midrule
\textbf{Merged} & \textbf{100885} & \textbf{230} & \textbf{1769} & \textbf{441} & \textbf{4355} & - \\
\bottomrule
\end{tabular}%
}
\captionsetup{skip=5pt}
\caption{Data Codes: BPM: BeetlePUUM, BPZ: BeetlePalooza, NHMC: NHM-Carabids. Column \{Genus/Species\} N/A: number of images where specimens are not identified to \{genus/species\} level; T+E Data: availability of morphological trait measurements (elytra length and width) and environment data.}
\label{tab:dataset_statistics}
\end{table}

% \paragraph{Data availability statement.} All datasets except BeetlePUUM, which is in preparation for publication, are publicly available.

% \begin{figure}[htbp]
%     \centering
%     \fbox{\includegraphics[width=0.22\columnwidth,height=4cm,keepaspectratio]{figures/puum.png}}\hfill
%     \fbox{\includegraphics[width=0.22\columnwidth,height=4cm,keepaspectratio]{figures/palooza.png}}\hfill
%     \fbox{\includegraphics[width=0.22\columnwidth,height=4cm,keepaspectratio]{figures/nhm.jpg}}\hfill
%     \fbox{\includegraphics[width=0.22\columnwidth,height=4cm,keepaspectratio]{figures/i1m.png}}
%     \captionsetup{skip=3pt}
%     \caption{Sample beetle specimen from four datasets. From left: BeetlePUUM, BeetlePalooza, NHM-Carabids, and 1M-Carabids}
%     \label{fig:beetles}
% \end{figure}

\subsection{Exploratory Data Analysis}
\label{subsec:analysis}
We start with an exploratory data analysis. Dataset overlap shows minimal taxonomic sharing between datasets, with I1MC having some overlap with both NHM-Carabids and BeetlePalooza; and BeetlePUUM being isolated due to its endemic focus (details in Table~\ref{tab:dataset_overlap}.)
\begin{table}[t]
\centering
\setlength{\tabcolsep}{7pt}
\resizebox{0.8\columnwidth}{!}{%
\begin{tabular}{l*{5}{c}}
    \toprule
      & I1MC & BPM & NHMC & BPZ \\
    \midrule
    \midrule
    I1MC & 206\textbackslash1531 & 2  & 68  & 73  \\
    BPM  & 2  & 4\textbackslash14  & 2  & 0   \\
    NHMC & 57 & 2  & 77\textbackslash290  & 2   \\
    BPZ  & 36 & 1  & 16  & 36\textbackslash76 \\
    \bottomrule
\end{tabular}
}
\captionsetup{skip=5pt}
\caption{Overlapping taxa across dataset. Upper triangle: Number of common species; Lower triangle: Number of common genera. Diagonal: Number of \{genera\textbackslash species\} in that dataset}
\label{tab:dataset_overlap}
\end{table}
Sample distribution per species varies significantly across datasets. NHM-Carabids demonstrates a balanced distribution with abundant samples per species. I1MC, in constrast, has a highly skewed distribution: nearly half of its species are rarely sampled ($\le$3 images). BeetlePUUM covers fewer species but with dense sampling, whereas BeetlePalooza offers moderate diversity but skewed representation. (Details in Appendix~\ref{app_subsec:data-analysis})

\begin{figure*}[!t]
\centering
\fbox{\includegraphics[width=\textwidth]{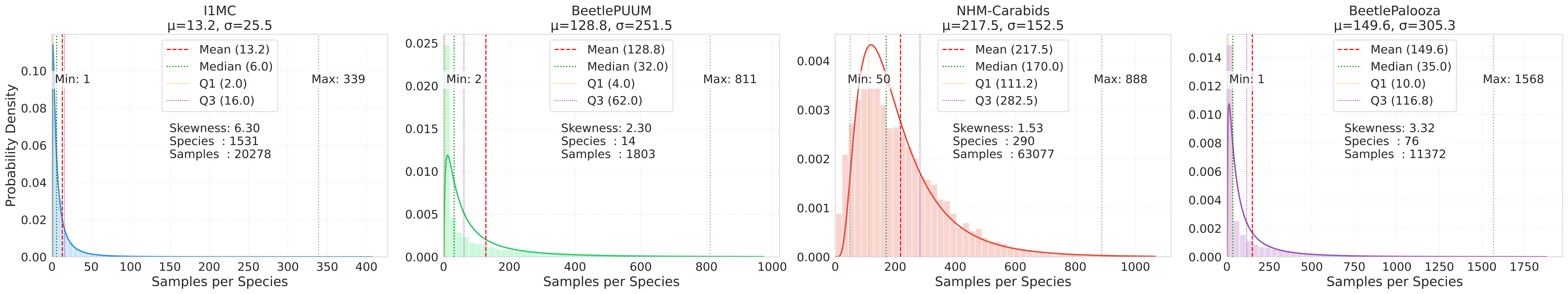}}
\captionsetup{skip=3pt}
\caption{Probability density distributions of datasets. X-axis: Number of samples per species (abundance). Y-axis: Probability density (relative frequency of species with given sample count). Histogram: Distribution of species by number of samples. Curve: Fitted probability distribution based on dataset statistics. The plots illustrate the variation in samples per species, with fitted probability distributions and key statistical parameters including mean, median, and quartile ranges (Q1, Q3). All four datasets exhibit characteristic right-skewed distributions (skewness values from 1.53 to 6.30), reflecting the common long-tailed pattern in ecological datasets where a few species are extensively sampled while most are represented by relatively few specimens.}
\label{fig:edahist}
\end{figure*}

\subsection{Taxonomic Classification}
\label{subsec:benchmarking}
Our primary focus is to evaluate the performance of state-of-the-art models on the taxonomic classification task across all datasets. To achieve this, we extract feature embeddings from various pretrained foundation models and train a multi-layer perceptron (MLP) classifier on top of these embeddings. The embeddings are derived from diverse vision architectures, categorized into: (1) vision-only self-supervised models (DINOv2 \cite{oquab2023dinov2}, ViTMAE \cite{he2022masked}, SwAV \cite{goyal2021self}, MoCov3 \cite{chen2021empirical}), (2) vision-only supervised models (SWINv2 \cite{liu2022swin}, BeIT \cite{bao2021beit}, LeViT \cite{graham2021levit}, ConvNeXt \cite{liu2022convnet}), and (3) vision+language supervised models (CLIP \cite{radford2021learning}, BioCLIP \cite{stevens2024bioclip}, SigLIP \cite{zhai2023sigmoid}, ViLT \cite{kim2021vilt}). For example, the carabid \textit{Carabus vietinghoffii} belongs to the family Carabidae, genus \textit{Carabus}, and species \textit{vietinghoffii}, illustrating the hierarchical structure used in our classification. As a zero-shot baseline, we use BioCLIP \cite{stevens2024bioclip}, a CLIP-variant for fine-grained biological taxonomic classification.

\subsection{Sample Efficient Probing}
\label{subsec:efficient_probing}
Next we investigate the critical question, \textit{how many labeled examples per taxon are required to achieve satisfactory results?} Using the NHM-Carabids dataset, we construct training subsets of increasing size through two sampling strategies: 1) Balanced Sampling, with K (10, 20, 50) samples per species; 2) Proportional Sampling, where class distributions are preserved while matching the  number of images in balanced sets (Details in Appendix~\ref{app_subsubsec:probing}). For both setups, we further compare performance using 50\% (Half-Set) and 100\% (Full-Set) of the available data, evaluating the top-performing vision models across these configurations. \label{app_subsubsec:probing}

% \subsection*{Task 2 : Efficient Recognition with Limited Data}
% \label{subsec:efficient_probing}

% Following our benchmarking experiments, we investigated a critical question for practical deployment: \textit{How many labeled examples per taxon are required to achieve satisfactory identification performance?}

% Using the NHM-Carabids dataset (our largest collection with approximately 63,000 images), we constructed multiple training subsets of increasing size through two sampling strategies:

% \begin{itemize}
%     \item \textbf{Balanced Sampling}: We created subsets containing exactly K examples per species, with K = \{10, 20, 50\} specimens per species.
%     \item \textbf{Proportional Sampling}: We maintained the original class distribution while sampling the same total number of images as in the balanced sets.
% \end{itemize}
% For both cases, we also evaluated performance using 50\% (Half-Set) and 100\% (Full-Set) of the available training data, comparing the top performing vision models.

\subsection{Cross-Dataset Domain Adaptation}
\label{subsec}
In our third experiment, we assess another logistical constraint for biodiversity researchers: \textit{how well do pretrained vision models generalize across datasets?} We experiment in two scenarios: (1) lab to lab adaptation, where we train on NHM-Carabids and test on BeetlePalooza - both containing lab collections only; (2) lab to in-situ adaptation, where we train separately on NHM-Carabids and BeetlePalooza, and test on I1MC (that contains in-situ images). Evaluations are performed at both genus and species levels and included only taxa present in both source and target datasets to enable direct performance comparison across shared taxa.

\section{Results}
\label{sec:results}

\subsection{Model Architecture and Data Preparation}
\label{subsec:data_preparation}
For classification, we employed a simple single hidden-layer MLP architecture to evaluate the features extracted from our pretrained vision model. The data was split using an 80/20 train-test ratio. Images were transformed using the ImageNet mean \([0.485, 0.456, 0.406]\) and standard deviation \([0.229, 0.224, 0.225]\), and resized to \(224 \times 224\) pixels. Features were standardized to zero mean and unit variance using a standard scaler fitted on the training set to prevent information leakage between the training and test data.

\subsection{Taxonomic Classification}
\label{subsec:results_benchmarking}

\subsubsection{Zero-Shot Prediction with BioClip}
\label{subsec:results_zero-shot}
Our first evaluation focuses on the zero-shot prediction performance of BioCLIP across three taxonomic levels (Table~\ref{tab:bioclip_zero_shot_results}). Accuracy is highest at the family level and declines rapidly through the genus to the species level. This gradual drop is attributed to the model's limited exposure to diverse taxonomic groups, as BioCLIP’s training data lacks sufficient representation of carabid taxa. Consequently, zero-shot results for I1MC and NHM-Carabids are omitted due to their limited taxonomic overlap with BioCLIP’s training data, which reduces its applicability to these diverse datasets and is partially reflected in the challenges observed with the merged dataset.
\begin{table}[!ht]
\centering
\captionsetup{skip=3pt}
\setlength{\tabcolsep}{5pt}
\caption{Zero-Shot Taxonomic using BioClip}
\label{tab:bioclip_zero_shot_results}
\resizebox{0.7\columnwidth}{!}{%
\begin{tabular}{l*{4}{c}}
\toprule
\textbf{Dataset} & \textbf{Family} & \textbf{Genus} & \textbf{Species} \\
\midrule
\midrule
BeetlePUUM     & 98.15 & 55.43 & 23.49 \\
BeetlePalooza  & 89.95 & 36.21 & 11.64 \\
Merged Data   & 77.37 & 23.17 & 3.21 \\
\bottomrule
\end{tabular}%
}
\end{table}

\subsubsection{Vision Embedding and Probing based Prediction}
\label{subsec:results_probing}
We report our benchmark evaluations in Tables~\ref{tab:benchmark_results}, ~\ref{tab:benchmark_results_macro}, \ref{tab:benchmark_details} with micro-accuracy, macro-accuracy, and detailed scores. Results from Table~\ref{tab:benchmark_results} show that ViLT performs the best across all datasets at both taxonomic levels, with a micro-accuracy of 97.15\% at genus and 93.97\% at species level on the merged data. However, the macro accuracy (from Table ~\ref{tab:benchmark_results_macro}) is relatively low, with 78.30\% at genus and 65.67\% at species level, implying that the performance may be weaker for less common or rarer taxa. We observe similar patterns for other models as well. In vision-only categories, DINOv2 leads self-supervised and BeIT leads supervised models (See Appendix~\ref{app_subsubsec:benchmarking} for detailed performance report).\\
Compared to the previous work on NHM-Carabids which reported 51.9\% species-level and 74.9\% genus-level accuracy using CNN \cite{hansen2020species}, we achieve significantly higher performance with 99.5\% accuracy at the species level and 99.8\% at the genus level (see Table~\ref{tab:benchmark_details}).

% Compared to previous works, we achieve significantly superior performance. As shown in Table~\ref{tab:comparison_prev_works}, our approach yields the highest accuracy while covering a substantially greater number of genera and species.
% \begin{table}[!ht]
% \centering
% \setlength{\tabcolsep}{3pt}
% \resizebox{\columnwidth}{!}{%
% \begin{tabular}{l*{5}{c}}
% \toprule
% \textbf{Dataset} & \textbf{Images} & \textbf{Genera} & \textbf{Species} & \textbf{Model} & \textbf{Score (\%)} \\
% \midrule
% \midrule
% \cite{blair2020robust} & 3,270 & 32 & 64 & LDA & 84.7 \\
% \cite{hansen2020species} & 63,364 & 80 & 291 & CNN & 51.9 \\
% \cite{abeywardhana2021deep} & 1,215 & 9 & N/A & SqueezeNet & 91.3 \\
% \midrule
% \textbf{Merged} & \textbf{100,885} & \textbf{230} & \textbf{1,769} & \textbf{ViLT+MLP} & \textbf{93.9} \\
% \bottomrule
% \end{tabular}%
% }
% \captionsetup{skip=3pt}
% \caption{CHUCK: NO!!!!  NEVER MAKE EXPERIMENTAL COMPARISON CLAIMS UNLESS YOU HAVE THE SAME DATASET. OTHERWISE IT IS MEANINGLESS!!!  PLEASE REMOVE.   Comparison with previous works: \cite{blair2020robust} used LDA (Linear Discriminant Analysis on digitized NEON specimen; \cite{hansen2020species} used CNN (Convolutional Neural Network) on NHM-Carabids; \cite{ariznavarreta2024multimodal} used SqueeNet on a small set of Sri Lankan tiger beetles.}
% \label{tab:comparison_prev_works}
% \end{table}

\subsection{Sample Efficient Probing}
\label{subsec:results_efficient_probing}
Starting from the top performing model ViLT on the full-dataset taxonomic classification task, we systematically analyze its performance across reduced sized training subsets, ranging from 2,900 at the small end to 63,077 images with proportional sampling (See Appendix~\ref{app_subsubsec:probing} for more details). 
\begin{figure}[!b]
\centering
\includegraphics[width=\columnwidth]{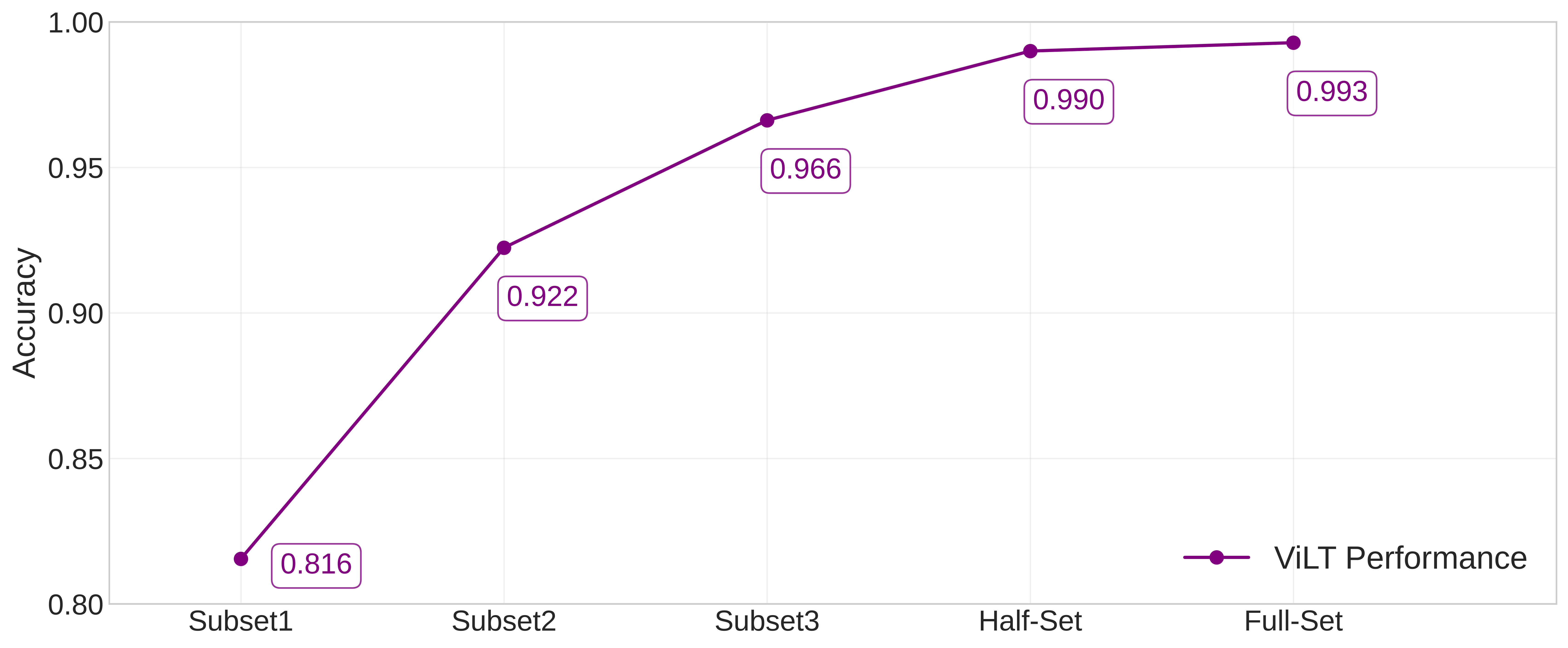}
\captionsetup{skip=3pt}
\caption{Performance of ViLT with different subsets of samples being probed. Subset1: 2900 images, Subset2: 5800 images, Subset3: 14500 images, Half-Set: 30000 images, Full-Set: 63077 images. Y-axis is the model's accuracy score on corresponding data.}
\label{fig:efficient_probing}
\end{figure}
We report the accuracy scores in Table~\ref{tab:probing_details}, that exhibit a pattern of diminishing returns: the most substantial performance improvement (+0.107) emerged between the first two subsets, with subsequent gains progressively attenuating and ultimately plateauing to a minimal +0.003 increment in the full data (See Figure~\ref{fig:efficient_probing}). This underlines a notable insight: more data does not always result in proportional improvements   , implying that strategically sampled data can achieve robust predictive capabilities while reducing computational resources.

\subsection{Cross-Dataset Domain Adaptation}
\label{subsec:results_domain_adaptation}
\begin{figure}[!ht]
\centering
\includegraphics[width=\columnwidth]{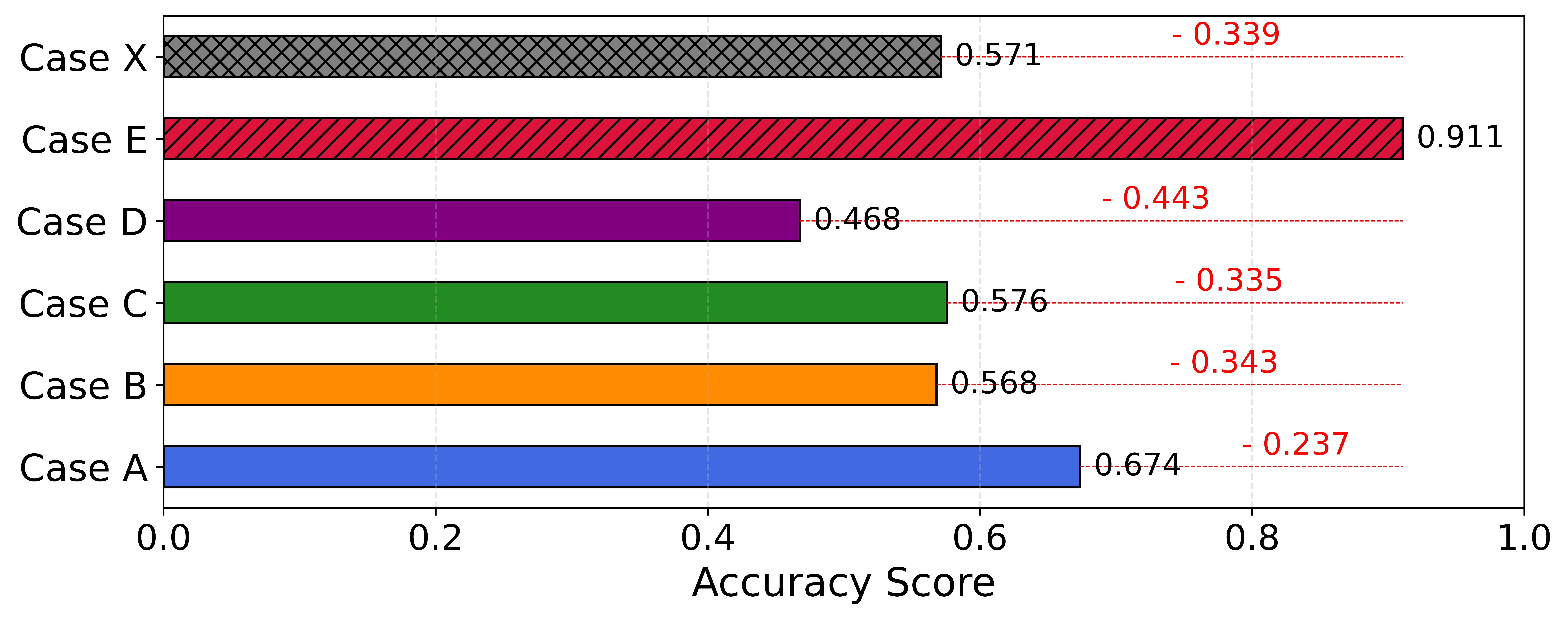}
\captionsetup{skip=2pt}
\caption{Cross-dataset domain adaptation performance of ViLT. Case A: Train on NHM-Carabids, Test on I1MC (at genus); Case B: same as A (at species); Case C: Train on BeetlePalooza, Test on I1MC (at genus); Case D: same as C (at species); Case E: Train on NHM-Carabids, Test on BeetlePalooza (at genus); Case X: Average of Cases A, B, C, D. Cases A to D represent lab-to-in-situ domain shifts for both genus and species levels, while Case E is evaluated at the genus level only due to limited species-level overlap between NHM-Carabids and BeetlePalooza.}
\label{fig:domain_adaptation}
\end{figure}

Figure~\ref{fig:domain_adaptation} illustrates a comparative overview of the performance of the top-performing model ViLT across two adaptation scenarios. The model generalizes well when trained on one lab dataset (NHM-Carabids) and tested on another (BeetlePalooza), achieving a score of 0.91 at the genus level, but performs poorly when tested on in-situ images (I1MC), with an average accuracy score of 0.57. This performance drop is driven by morphological and contextual variations, such as variable lighting, non-standard orientations (e.g., lateral views), heterogeneous substrates (e.g., leaf litter, soil textures), occlusions by debris or vegetation, moisture effects on cuticular reflectance, and the presence of microorganisms (e.g., mites, fungi) in I1MC, unlike the controlled dorsal views in NHM-Carabids. These scores highlight the persistent challenge of cross-domain generalization in taxonomic classification \cite{shi2024potential}. Detailed performance scores are reported in Table~\ref{tab:domain_shift}.
The poor performance on I1MC underscores the need for domain-adaptive strategies to address ecological variability and visual noise, enhancing model robustness across diverse domains.

% CHUCK: I PREDICT THIS WILL HURT US. WE SHOULD BE TRAINING ON ALL LAB DATASETS AND TESTING ON IN SITU.  I DOUBT THERE IS TIME TO FIX THIS, HOWEVER.

% CHUCK: ANOTHER THING THAT WOULD BE COOL WOULD BE A PLOT OF ACCURACY AS A FUNCTION OF THE  NUMBER OF TRAINING EXAMPLES. 

\section{Conclusion and Future Work}
\label{sec:conclusion}
% Our work presents a significant advancement in streamlining carabid taxonomy. However, there are critical gaps that need to be resolved. Firstly, although we compile data from four taxonomically distant sources, still it only covers $<$5\% of all carabid species. Second, the domain-shift score drop on I1MC data reveals a major domain gap. In the future, we will work on bridging these gaps, as well as analyzing predicted taxa using phylogenetic and ecological proximity, and testing the models ability to flag a novel taxa. With these efforts, we hope to develop a scalable framework of taxonomic classification for carabids, advancing ecological analysis and biodiversity monitoring.
% \textcolor{red}{grok: "While our study trains separate classifiers for genus and species levels, this approach may lead to inconsistent hierarchical predictions. Future work could explore hierarchical classification methods, such as Flamingo [1] or HAFeat [2], to ensure consistent taxonomic predictions by integrating the known taxonomy tree during training."
% \\
% "The limited taxonomic overlap between I1MC and other datasets (Table 2) highlights the potential for open-set recognition methods to address unseen classes, which we plan to investigate in future work."}
Our work presents a significant advancement in streamlining carabid taxonomy. However, critical gaps remain to be addressed. Firstly, although we compile data from four taxonomically diverse sources, these still cover less than 5\% of all carabid species. Secondly, the domain-shift score drop on I1MC data reveals a significant domain gap. While our study trains separate classifiers for genus and species levels, this approach may lead to inconsistent hierarchical predictions. To address the possibility of inconsistent hierarchical predictions for genus and species levels, we look forward to exploring hierarchical classification methods, such as Flamingo \cite{chang2021your}, or HAF \cite{garg2022learning}, to ensure consistent taxonomic predictions by integrating the known taxonomy tree during training. Additionally, the limited taxonomic overlap between I1MC and other datasets (Table~\ref{tab:dataset_overlap}) highlights the potential for open-set recognition methods to address unseen classes, which we plan to investigate in future work. Moving forward, we aim to bridge these gaps by analyzing predicted taxa using phylogenetic and ecological proximity and testing the models' ability to flag novel taxa. Through these efforts, we hope to develop a scalable framework for taxonomic classification of carabids, advancing ecological analysis and biodiversity monitoring.

{
    \small
    \bibliographystyle{ieeenat_fullname}
    \bibliography{main}
}

\section*{Acknowledgment}
This work was supported by both the Imageomics Institute and the AI and Biodiversity Change (ABC) Global Center. The Imageomics Institute is funded by the US National Science Foundation's Harnessing the Data Revolution (HDR) program under Award No. 2118240 (Imageomics: A New Frontier of Biological Information Powered by Knowledge-Guided Machine Learning). The ABC Global Center is funded by the US National Science Foundation under Award No. 2330423 and Natural Sciences and Engineering Research Council of Canada under Award No. 585136. This work draws on research funded by the Social Sciences and Humanities Research Council. S. R. and A. E. were additionally supported by the US National Science Foundation's Award No. 242918 (EPSCOR Research Fellows: NSF: Advancing National Ecological Observatory Network-Enabled Science and Workforce Development at the University of Maine with Artificial Intelligence) and by Hatch project Award \#MEO-022425 from the US Department of Agriculture’s National Institute of Food and Agriculture. This work draws on research supported by the National Ecological Observatory Network (NEON), a program sponsored by the National Science Foundation and operated under cooperative agreement by Battelle. This material uses specimens and/or samples collected as part of the NEON Program and is based in part upon work supported by NEON. Additionally, this material is based upon work supported by the National Science Foundation under Award Numbers 2301322, 1950364, and 1926569. This publication uses data generated via the \href{https://www.zooniverse.org/}{Zooniverse} platform, the development of which is funded by generous support, including a Global Impact Award from Google, and by a grant from the Alfred P. Sloan Foundation.

% \paragraph{Data Availability Statement.} All four datasets are publicly available: BeetlePUUM (Hawaii-Beetles*) \cite{rayeed2025HawaiiBeetles}, BeetlePalooza (2018-NEON-Beetles*) \cite{Fluck2018_NEON_Beetle}, NHM-Carabids \cite{hansen_2019_3549369}, I1MC \cite{nguyen2024insect}.

\paragraph{Data Availability Statement.} All four datasets used in this study are publicly available: BeetlePUUM\footnote{Official name on Hugging Face: \href{https://huggingface.co/datasets/imageomics/Hawaii-beetles}{imageomics/Hawaii-beetles}} \cite{rayeed2025HawaiiBeetles}, BeetlePalooza\footnote{Official name on Hugging Face: \href{https://huggingface.co/datasets/imageomics/2018-NEON-beetles}{imageomics/2018-NEON-beetles}} \cite{Fluck2018_NEON_Beetle}, NHM-Carabids \cite{hansen_2019_3549369}, and I1MC\footnote{Official name: \href{https://uark-my.sharepoint.com/:f:/g/personal/hn016_uark_edu/Ekg5PAR-5GJKhdd3YTj2kBoBc0mNOcCP5YQqEnxjY4h-qg?e=dnBs5m}{uark-cviu.github.io/projects/insect-foundation}} \cite{nguyen2024insect}.

\paragraph{Code Availability Statement.} All code developed for and utilized in this study is publicly available on github at \href{https://github.com/Imageomics/BeetleVerse}{Imageomics/BeetleVerse}.

\paragraph{Competing Interests.} The authors declare no competing interests.

% \clearpage

\appendix

\newpage

\section*{Appendices}

\begin{enumerate}
    \item{\cref{app:method-details}: Methodology details for Section~\ref{sec:methodology}}
    \item{\cref{app:results-details}: Result details for Section~\ref{sec:results}}
    \item{\cref{app_sec:multimodality}: Incorporating Multimodal Data}
    \item{\cref{app_subsec:feature-mapping}: Feature Mapping for better Visualization}   
\end{enumerate}

\section{Methodology Details}
\label{app:method-details}

\subsection{Dataset Details}
\label{app_subsec:data-details}
In this study, we use four datasets, each contributing unique strengths in taxonomic coverage, imaging methodology, and ecological context. Below, we provide a comprehensive breakdown of each, including their curation processes, taxonomic scope, and key characteristics. 

\paragraph{BeetlePUUM.}
This dataset digitizes all pinned carabid specimens from the NEON Pu'u Maka'ala ecological observatory site in Hawaii \cite{rayeed2025HawaiiBeetles}. The specimens originate from the NEON biorepository carabid pinned voucher collection curated by the Bernice Pauahi Bishop Museum \cite{NEON-pinned-specimens}, with associated specimen- and trap-level metadata provided through the NEON ground beetle pitfall sampling data product \cite{NEON-pinned-beetles-metadata}. The dataset is publicly available on HuggingFace, at \href{https://huggingface.co/datasets/imageomics/Hawaii-beetles}{imageomics/Hawaii-beetles}. The collection was assembled from 600 original source images: 420 bulk images captured with a Canon EOS DSLR camera (model 7D with a 24--105 macro lens) and 180 high-detail microscopic images acquired using a SWIFCAM SC1603 system featuring a 16MP 1/2.33" CMOS sensor. The bulk images each contained 3--5 pinned specimens arranged vertically in sequential order according to their unique specimen IDs, with all individuals within an image representing the same species and captured from the same pitfall trap. 

% To ensure the images are optimized for advanced ecological applications such as automated trait extraction, we follow the digitization guidelines outlined in \cite{east2025optimizing}, which emphasize standardized specimen positioning, consistent size and color calibration, and comprehensive metadata documentation. We then apply Grounding DINO \cite{liu2024grounding} to precisely detect and crop individual beetles from the group images. The original group images include comprehensive metadata comprising geolocation coordinates, collection dates, and taxonomic authentication by carabid specialists. Using the collection coordinates and dates, we extract relevant weather data for each specimen to provide ecological context for morphological analyses. Morphological traits are measured using the TORAS digital annotation tool \cite{torontoannotsuite}. Figure~\ref{fig:puum_sample} displays a representative group image and the corresponding individual specimen crops.

\begin{figure}[!ht]
  \centering
  \includegraphics[width=\columnwidth]{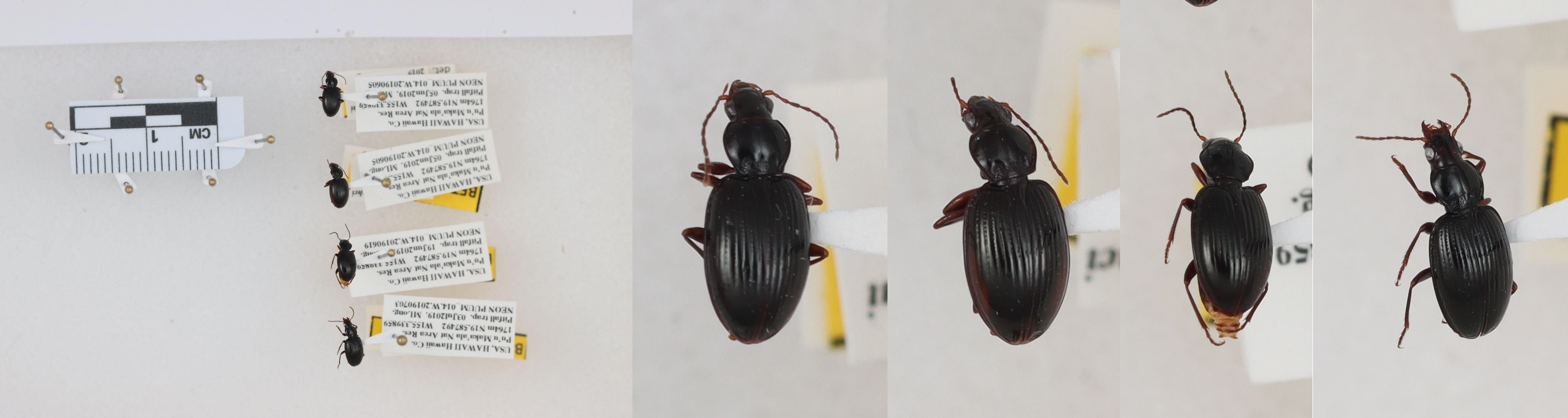}
   \captionsetup{skip=2pt}
   \caption{A sample group image and corresponding individual crops from the BeetlePUUM dataset. Leftmost panel shows the group image with measurement scale, while the four right panels present images of those specimens individually cropped.}
   \label{fig:puum_sample}
\end{figure}

\noindent To ensure the images are optimized for advanced ecological applications such as automated trait extraction, the curators of the dataset follow the digitization guidelines outlined in \cite{east2025optimizing}, which emphasize standardized specimen positioning, consistent size and color calibration, and comprehensive metadata documentation. Grounding DINO \cite{liu2024grounding} is then applied to precisely detect and crop individual beetles from the group images. The original group images include comprehensive metadata comprising geolocation coordinates, collection dates, and taxonomic authentication by carabid specialists. Using the collection coordinates and dates, relevant weather data are extracted for each specimen to provide ecological context for morphological analyses. Morphological traits are measured using the TORAS digital annotation tool \cite{torontoannotsuite}. Figure~\ref{fig:puum_sample} displays a representative group image and the corresponding individual specimen crops.

\paragraph{BeetlePalooza.}
This dataset digitizes NEON carabid specimens and substantially expands both geographic and taxonomic coverage, comprising 11,399 images collected from 30 NEON sites across the continental United States \cite{Fluck2018_NEON_Beetle}. The specimens are drawn from the NEON biorepository ethanol-preserved collections, associated with trap sorting and archive pooling workflows \cite{Portal2022-ho, Portal2022-qu}. Unlike traditional pinned voucher specimens, this dataset focuses on excess individuals preserved in ethanol-filled vials — specifically, specimens collected beyond ten individuals per pitfall trap, which are pinned separately under NEON protocols. The dataset is publicly available on HuggingFace, at \href{https://huggingface.co/datasets/imageomics/2018-NEON-beetles}{imageomics/2018-NEON-beetles}.

\begin{figure}[!ht]
  \centering
  \fbox{\includegraphics[width=0.95\columnwidth]{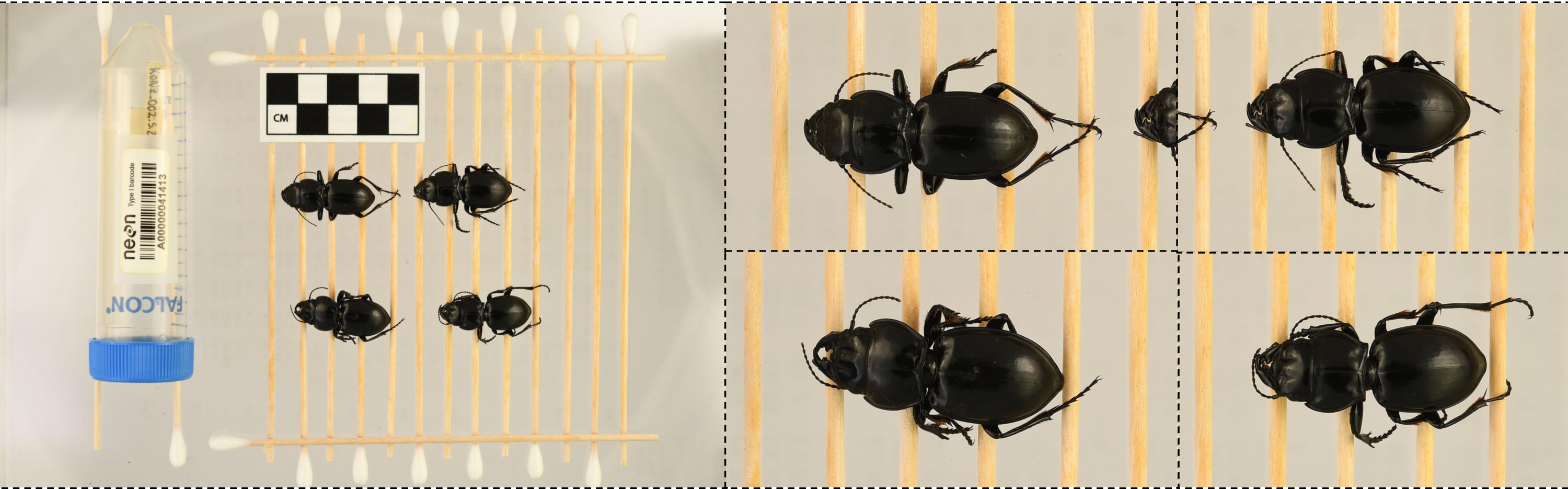}}
   \captionsetup{skip=2pt}
   \caption{A sample group image and corresponding individual crops from the BeetlePalooza dataset. Leftmost panel shows the group image with measurement checkbox, while the right panels present individual crops of the same specimens.}
   \label{fig:palooza_sample}
\end{figure}

\noindent During digitization, specimens are carefully air-dried to remove residual ethanol, mounted on minute staging sticks to standardize orientation, and imaged in bulk. Due to the delicate nature of ethanol-preserved specimens, some individuals could not be repositioned, resulting in variability in specimen orientation. As with BeetlePUUM, Grounding DINO \cite{liu2024grounding} is applied to isolate individual beetles from group images. Each specimen image is linked with NEON-provided collection metadata and supplemented with site-level weather data derived from collection date and geolocation. Morphological traits for this dataset are annotated using the Zooniverse digital annotation platform \cite{simpson2014zooniverse}. Figure~\ref{fig:palooza_sample} presents a representative group image and the corresponding individual specimen crops produced through the detection and cropping pipeline.

\noindent \cite{rayeed2026continental} presents a compilation of these two datasets, featuring meticulously-measured morphological traits of individual specimens and offering a trait-based foundation for exploring taxonomic relationships and ecological variation among carabids. Moreover, it serves as a valuable testbed for small-data regimes with multimodal data, a relatively less-explored area in ML where conventional approaches often underperform \cite{todman2023small, stevens2025mind}.

\paragraph{NHM-Carabids.}
This dataset has 63,077 high-resolution habitus images of 361 carabid species from the British Isles, digitized from the curated collections of Natural History Museum in London. All specimens are taxonomically verified to species level, with metadata including collection dates (spanning 150+ years), collector annotations, and morphological descriptors. Imaging was performed under controlled lighting, though some historical digitizations exhibit moderate blurring. 
\begin{figure}[!ht]
  \centering
  \includegraphics[width=\columnwidth]{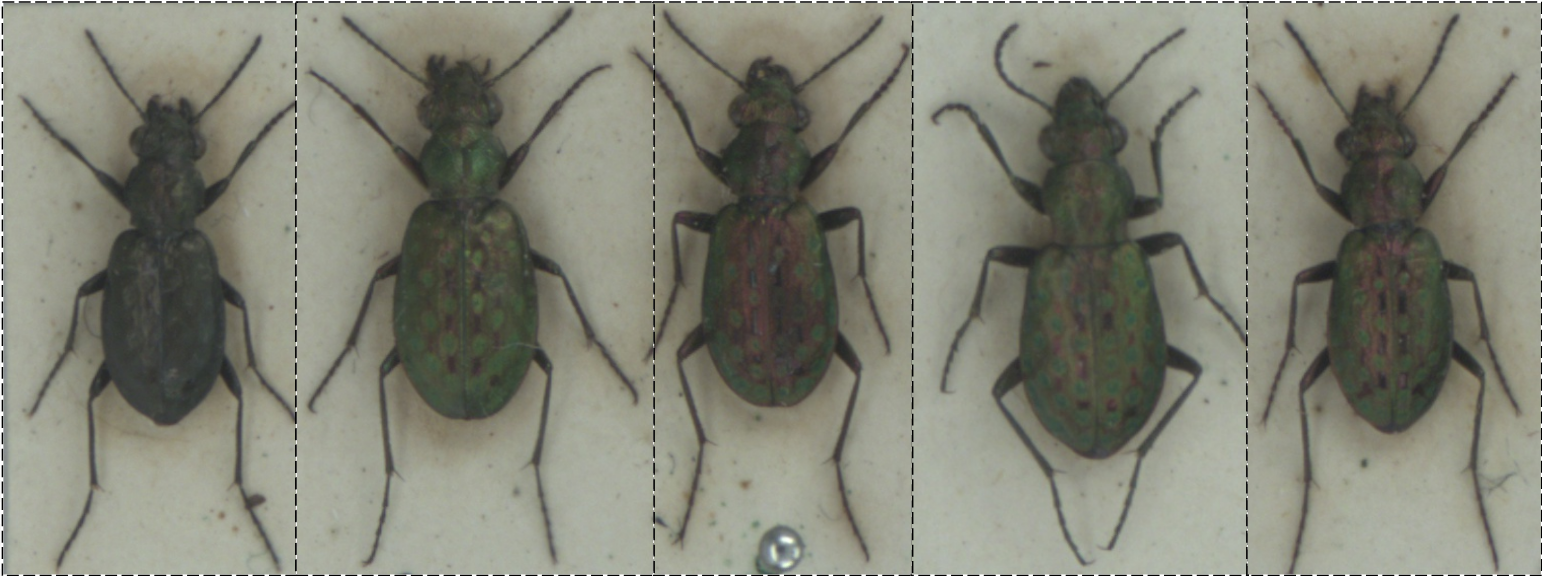}
   \captionsetup{skip=2pt}
   \caption{Sample specimens from the NHM-Carabids dataset}
   \label{fig:NHM_sample}
\end{figure}
As a museum collection, it lacks ecological metadata but provides an unparalleled reference for alpha taxonomy, temporal trait shifts, and rare specimen studies. Experiments on species-level classification with CNN revealed that larger-bodied species and those in less speciose genera were classified more reliably. Figure ~\ref{fig:NHM_sample} illustrates a few sample museum specimens.

\begin{figure}[!ht]
  \centering
  \includegraphics[width=\columnwidth]{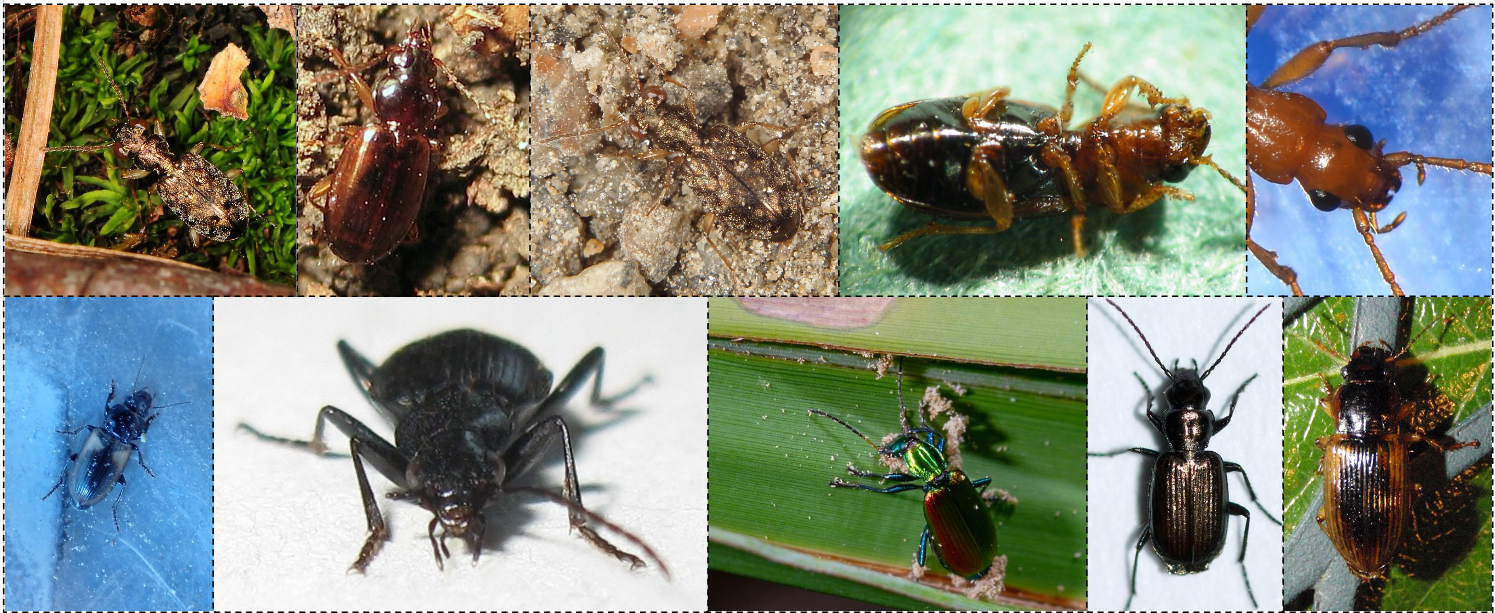}
   \captionsetup{skip=2pt}
   \caption{Sample specimens from the I1MC dataset. Complex backgrounds: Top row, 1st to 3rd images; Varying viewpoints: Ventral (Top row, 4th image), Anterior (Bottom row, 2nd image); Partial occlusion: Bottom row, 3rd image; Incomplete: Top row, 5th image; Lighting/Shadow: Bottom row, 5th image}
   \label{fig:I1M_sample}
\end{figure}

\paragraph{I1MC.} 
Diverse but Noisy Field and Lab Imagery
Extracted from the Insect-1M foundational dataset, this subset includes 24,606 carabid images combining in-situ field observations and lab-digitized specimens. Sourced from naturalist-contributed HTML repositories, the raw data underwent expert vetting to remove mislabeled, corrupted, or non-insect images, resulting in a cleaned dataset with hierarchical taxonomic labels (Subphylum to Species). Despite covering 206 genera and 1,531 species, the broadest taxonomic range among all datasets used, only 60\% of images are identified to species level due to community-sourced limitations (4328 samples not identified to species level; and 424 samples not identified to genus level). Field images often exhibit occlusions or uneven lighting, while lab specimens vary in preservation quality. The dataset’s strength lies in its ecological context and scale, supporting pretraining for generalist vision models. On the other hand, its primary limitation stems from image quality inconsistencies: frequently having images of parts of a beetle rather than a full beetle pictured, varying perspective orientations (dorsal, lateral, ventral, and anterior), and recurrent issues with image clarity and focus.
Figure~\ref{fig:I1M_sample} illustrates some examples of these.

\subsection{Exploratory Data Analysis}
\label{app_subsec:data-analysis}
We conduct an exploratory data analysis on the datasets to uncover patterns in taxonomic diversity, sample distribution, and dataset overlap. The analysis leverages summary statistics, quartile distributions, abundance classifications, Jaccard indices for overlap, and visualizations of sample distributions to provide a comprehensive understanding of the datasets.

\subsubsection{Summary Statistics and Distributional Insights}

\begin{table*}[!t]
\centering
\resizebox{\textwidth}{!}{%
\begin{tabular}{lrrrrrrrrrrrr}
    \toprule
    Dataset & Mean & Median & Std Dev & Min & Q1 (25\%) & Q3 (75\%) & IQR & Max & Total Genera & Total Samples & Skewness & Kurtosis \\
    \midrule
    \midrule
    I1MC & 117.39 & 25.5 & 269.17 & 1 & 8.00 & 83.50 & 75.50 & 2457 & 206 & 24182 & 4.93 & 32.22 \\
    BeetlePUUM & 450.75 & 333.5 & 523.45 & 9 & 52.50 & 731.75 & 679.25 & 1127 & 4 & 1803 & 0.79 & -1.54 \\
    NHM-Carabids & 819.18 & 355.0 & 1715.11 & 50 & 125.00 & 697.00 & 572.00 & 13298 & 77 & 63077 & 5.58 & 37.69 \\
    BeetlePalooza & 316.17 & 60.5 & 562.79 & 1 & 9.25 & 328.75 & 319.50 & 2242 & 36 & 11382 & 2.38 & 5.14 \\
    Merged & 438.62 & 58.0 & 1319.80 & 1 & 11.00 & 328.00 & 317.00 & 14771 & 229 & 100444 & 7.21 & 66.57 \\
    \addlinespace
    \hdashline
    \addlinespace
    Dataset & Mean & Median & Std Dev & Min & Q1 (25\%) & Q3 (75\%) & IQR & Max & Total Species & Total Samples & Skewness & Kurtosis \\
    \midrule
    I1MC & 13.24 & 6.0 & 25.52 & 1 & 2.00 & 16.00 & 14.00 & 339 & 1531 & 20278 & 6.30 & 53.01 \\
    BeetlePUUM & 128.79 & 32.0 & 251.45 & 2 & 4.00 & 62.00 & 58.00 & 811 & 14 & 1803 & 2.30 & 4.35 \\
    NHM-Carabids & 217.51 & 170.0 & 152.50 & 50 & 111.25 & 282.50 & 171.25 & 888 & 290 & 63077 & 1.53 & 2.83 \\
    BeetlePalooza & 149.63 & 35.0 & 305.27 & 1 & 10.00 & 116.75 & 106.75 & 1568 & 76 & 11372 & 3.32 & 11.72 \\
    Merged & 54.57 & 9.0 & 126.57 & 1 & 3.00 & 25.00 & 22.00 & 1581 & 1769 & 96530 & 4.96 & 36.83 \\
    \bottomrule
\end{tabular}
}
\captionsetup{skip=3pt}
\caption{Summary statistics for the datasets, divided into two sections. The top section (above the dashed line) presents descriptive statistics for genera, including mean, median, standard deviation, minimum, 1st quartile (Q1, 25\%), 3rd quartile (Q3, 75\%), interquartile range (IQR), maximum, total number of genera, total number of samples, skewness, and kurtosis. The bottom section (below the dashed line) provides the same statistical measures for species across the same datasets, with the total number of species replacing total genera.}
\label{tab:beetle_stats}
\end{table*}
\begin{figure*}[!t]
\centering
\fbox{\includegraphics[width=0.98\textwidth]{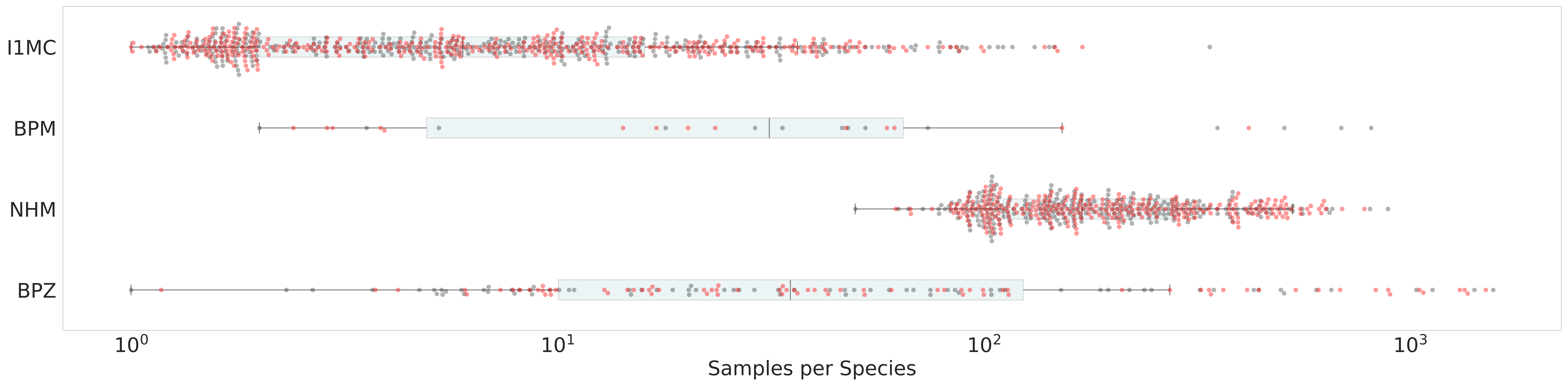}}
\captionsetup{skip=3pt}
% \caption{Sample distribution across datasets: X-axis depicts samples per species on a logarithmic scale, showing the distinct characteristics of each dataset. Scatter points overlaid on box plots illustrate the long-tailed distribution within datasets, particularly pronounced in I1MC.}
\caption{Horizontal swarm plot illustrating sample distribution across datasets: Data Codes are same as Table~\ref{tab:dataset_statistics}. X-axis: Number of samples per species on a logarithmic scale ($10^0$ to $10^3$). The boxplot shows interquartile range and median for each dataset, overlaid with a swarm plot, where each data point reflects the number of samples for a species. The plot highlights sampling disparities: NHM exhibits the most balanced distribution, with a relatively even spread of samples per species from $10^2$ to $10^3$, while I1MC shows a heavy skew toward minimal samples (near $10^0$), indicating significant undersampling. BPM and BPZ display significant variability, with some species having up to $10^2$ samples but still showing a skew toward lower sample counts.}
\label{fig:edaswarm}
\end{figure*}
Table~\ref{tab:beetle_stats} presents detailed summary statistics for both genera and species across the datasets. For genera, NHM-Carabids exhibits the highest mean samples per genus (819.18) and the largest maximum (13,298), but also the highest variability (standard deviation of 1,715.11), indicating a wide range of sampling efforts. In contrast, BeetlePUUM has a mean of 450.75 samples per genus but a much smaller total genera count (4), reflecting its focused scope. I1MC and BeetlePalooza show more moderate means (117.39 and 316.17, respectively), but both display high skewness (4.93 and 2.38) and kurtosis (32.22 and 5.14), suggesting long-tailed distributions with many genera having few samples and a few genera being heavily sampled. The merged dataset, combining all four, has a mean of 438.62 samples per genus but an extremely high skewness (7.21) and kurtosis (66.57), reflecting the combined effect of these skewed distributions. For species, the patterns shift. NHM-Carabids again shows a high mean (217.51 samples per species) with a relatively low standard deviation (152.50), indicating a more balanced distribution. BeetlePUUM, despite its small species count (14), has a mean of 128.79 samples per species, suggesting dense sampling within its limited scope. I1MC, however, has a low mean (13.24) and median (6.0), with a high skewness (6.30) and kurtosis (53.01), indicating that most species are sparsely sampled. BeetlePalooza shows a mean of 149.63 but a high maximum (1,568), reflecting a skewed distribution (skewness: 3.32, kurtosis: 11.72). The merged dataset for species has a mean of 54.57, with a median of 9.0, further emphasizing the prevalence of sparsely sampled species across the combined data.

\subsubsection{Quartile Distribution}
Table~\ref{tab:beetle_quantiles} provides quartile distributions for both species and genera, illustrating the spread and central tendencies. For species, NHM-Carabids stands out with a median (Q2) of 170.0 and a third quartile (Q3) of 282.50, reflecting a higher baseline of samples per species. BeetlePUUM, despite its small species count, has a median of 32.0 and a Q3 of 62.0, indicating dense sampling. In contrast, I1MC has a median of 6.0 and a Q3 of 16.0, showing that 75\% of its species have 16 or fewer samples. BeetlePalooza’s median is 35.0, but its Q4 (maximum) reaches 1,568, highlighting a long tail. For genera, NHM-Carabids again shows a high median (355.0) and Q3 (697.0), while BeetlePUUM’s median is 333.5, reflecting its focused but well-sampled genera. I1MC and BeetlePalooza have medians of 25.5 and 60.5, respectively, with maximum values (2,457 and 2,242) indicating the presence of a few heavily sampled genera.
\begin{table}[!t]
\centering
\resizebox{0.98\columnwidth}{!}{%
\begin{tabular}{lrrrrr}
    \toprule
    \multicolumn{1}{c}{} & \multicolumn{5}{c}{Quartile Distribution: Species} \\
    \cmidrule(lr){2-6}
    Dataset & Q0 (0\%) & Q1 (25\%) & Q2 (50\%) & Q3 (75\%) & Q4 (100\%) \\
    \midrule
    I1MC & 1.0 & 2.00 & 6.0 & 16.00 & 339.0 \\
    BeetlePUUM & 2.0 & 4.00 & 32.0 & 62.00 & 811.0 \\
    NHM-Carabids & 50.0 & 111.25 & 170.0 & 282.50 & 888.0 \\
    BeetlePalooza & 1.0 & 10.00 & 35.0 & 116.75 & 1568.0 \\
    
    \addlinespace
    \hdashline
    \addlinespace
    
    \multicolumn{1}{c}{} & \multicolumn{5}{c}{Quartile Distribution: Genera} \\
    \cmidrule(lr){2-6}
    Dataset & Q0 (0\%) & Q1 (25\%) & Q2 (50\%) & Q3 (75\%) & Q4 (100\%) \\
    \midrule
    I1MC & 1.0 & 8.00 & 25.5 & 83.50 & 2457.0 \\
    BeetlePUUM & 9.0 & 52.50 & 333.5 & 731.75 & 1127.0 \\
    NHM-Carabids & 50.0 & 125.00 & 355.0 & 697.00 & 13298.0 \\
    BeetlePalooza & 1.0 & 9.25 & 60.5 & 328.75 & 2242.0 \\
    \bottomrule
\end{tabular}
}
\captionsetup{skip=3pt}
\caption{Quartile distribution statistics for the datasets: The top section displays the quartile distribution for species, including the minimum (Q0, 0\%), first quartile (Q1, 25\%), median (Q2, 50\%), third quartile (Q3, 75\%), and maximum (Q4, 100\%) values. The bottom section provides the same quartile measures for genera across the same datasets. These statistics illustrate the spread and central tendencies of species and genera within each dataset.}
\label{tab:beetle_quantiles}
\end{table}

\subsubsection{Sample Distribution Visualization}
Figure~\ref{fig:edaswarm} visualizes the sample distribution per species on a logarithmic scale. I1MC shows a median of 6.0 and a mean of 13.2, with a total of 1,531 species, but its distribution is heavily skewed, with many species having fewer than 10 samples and a few outliers reaching up to 339. BeetlePUUM, with only 14 species, has a median of 32.0 and a mean of 128.8, indicating denser sampling, though its maximum is 811. NHM-Carabids, with 290 species, has a median of 170.0 and a mean of 217.5, showing a more balanced distribution, though outliers extend to 888. BeetlePalooza’s 76 species have a median of 35.0 and a mean of 149.6, with a maximum of 1,568, reflecting a skewed distribution. Figure~\ref{fig:edahist} illustrates these statistics on a logarithmic scale, with probability density curves. I1MC’s distribution is highly right-skewed (skewness: 6.30), with a peak near the lower end (1–10 samples) and a long tail extending to 339. BeetlePUUM’s distribution, despite its small species count, shows a peak around 32 samples but extends to 811, with a skewness of 2.30. NHM-Carabids has a more symmetric distribution (skewness: 1.53), peaking around 170 samples, though it still has a tail up to 888. BeetlePalooza’s distribution is skewed, with a peak near 35 samples and a long tail reaching 1,568.

\subsubsection{Species Abundance Classification}
Table~\ref{tab:beetle_abundance} classifies species into four abundance categories: Rare, Uncommon, Common, and Abundant. I1MC has a striking 48.99\% of its species (750) classified as Rare, and 30.63\% (469) as Uncommon, with only 1.70\% (26) being Abundant, confirming its highly skewed distribution. BeetlePUUM, with only 14 species, has 35.71\% (5) Rare and 21.43\% (3) Abundant, reflecting its dense sampling within a small scope. NHM-Carabids has no Rare or Uncommon species, with 21.03\% (61) Common and 78.97\% (229) Abundant, highlighting its balanced and abundant sampling. BeetlePalooza shows a more even spread, with 18.42\% (14) Rare, 21.05\% (16) Uncommon, 32.89\% (25) Common, and 27.63\% (21) Abundant, indicating moderate diversity but skewed representation.
\begin{table}[!t]
\centering
\resizebox{\columnwidth}{!}{%
\begin{tabular}{lrrrrr}
    \toprule
    Dataset Codes & Rare & Uncommon & Common & Abundant \\
    \midrule
    I1MC & 750 (48.99\%) & 469 (30.63\%) & 286 (18.68\%) & 26 (1.70\%) \\
    BPM & 5 (35.71\%) & 1 (7.14\%) & 5 (35.71\%) & 3 (21.43\%) \\
    NHM & 0 (0.00\%) & 0 (0.00\%) & 61 (21.03\%) & 229 (78.97\%) \\
    BPZ & 14 (18.42\%) & 16 (21.05\%) & 25 (32.89\%) & 21 (27.63\%) \\
    \bottomrule
\end{tabular}
}
\caption{Species abundance classification: The table categorizes species into four abundance classes based on their counts: Rare (less than 5), Uncommon (5–20), Common (21–100), and Abundant (more than 100). For each dataset, the number of species in each category is shown, followed by the percentage of total species in that dataset. This classification highlights the distribution of species abundance, reflecting differences in rarity and prevalence across the datasets. Dataset codes are same as Table ~\ref{tab:dataset_statistics}.}
\label{tab:beetle_abundance}
\end{table}

\subsubsection{Taxonomic Overlap}
Taxonomic overlap between datasets was assessed using the Jaccard index\footnote{The Jaccard Index between two sets \( A \) and \( B \) is defined as:
\( J(A, B) = \frac{|A \cap B|}{|A \cup B|} \) where \( |A \cap B| \) is the size of the intersection of sets \( A \) and \( B \), and \( |A \cup B| \) is the size of their union. If \( A \) and \( B \) are empty, \( J(A, B) \) is defined as 1.} (Table~\ref{tab:jaccard_overlap}) and raw counts of common taxa (Table~\ref{tab:dataset_overlap}). The Jaccard index reveals minimal overlap overall. For species (upper triangle), I1MC shares the highest overlap with BeetlePalooza (0.0469) and NHM-Carabids (0.0388), while BeetlePUUM (BPM) shows very low overlap with all datasets (e.g., 0.0013 with I1MC). For genera (lower triangle), NHM-Carabids and I1MC have the highest overlap (0.2522), followed by BeetlePalooza and I1MC (0.1691). BeetlePUUM remains isolated, with overlaps as low as 0.0096 with I1MC. Table~\ref{tab:dataset_overlap} provides raw counts: I1MC shares 68 species with NHM-Carabids and 73 with BeetlePalooza, while BeetlePUUM shares only 2 species with I1MC and NHM-Carabids, and none with BeetlePalooza. For genera, I1MC and NHM-Carabids share 57 genera, while BeetlePUUM shares only 2 genera with I1MC and NHM-Carabids, and 1 with BeetlePalooza. This confirms BeetlePUUM’s isolation, likely due to its endemic focus, while I1MC shows moderate overlap with NHM-Carabids and BeetlePalooza.
\begin{table}[!t]
\centering
\setlength{\tabcolsep}{7pt}
\resizebox{0.7\columnwidth}{!}{%
\begin{tabular}{l*{4}{c}}
    \toprule
      & I1MC & BPM & NHM & BPZ \\
    \midrule
    I1MC & -      & 0.0013   & 0.0388     & 0.0469    \\
    BPM & 0.0096 & -      & 0.0253     & 0.0256    \\
    NHM & 0.2522 & 0.0253 & -          & 0.1649    \\
    BPZ & 0.1691 & 0.0256 & 0.1649     & -         \\
    \bottomrule
\end{tabular}
}
\captionsetup{skip=5pt}
\caption{Jaccard index values representing the overlap of genera and species: The upper triangle indicates the Jaccard index for the number of common species shared between pairs of datasets, while the lower triangle represents the Jaccard index for the number of common genera. Values range from 0 (no overlap) to 1 (complete overlap), with higher values indicating greater similarity. Dataset codes correspond to those defined in Table~\ref{tab:dataset_statistics}.}
\label{tab:jaccard_overlap}
\end{table}

\begin{figure*}[!t]
    \centering
    \includegraphics[width=\textwidth]{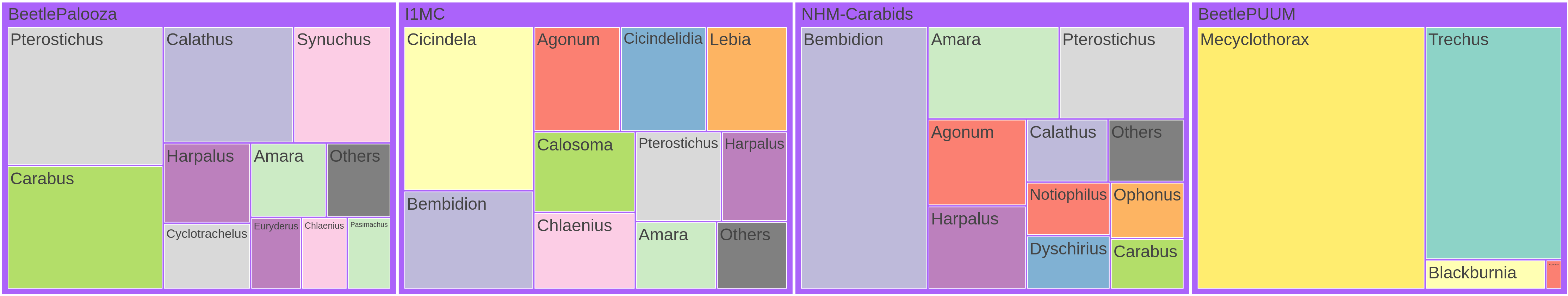}
    \captionsetup{skip=2pt}
    \caption{Treemap representation of genus distribution across four datasets. In each dataset, the top 10 genera by frequency are shown individually, with all other genera combined into an `Others' category. Rectangle sizes are normalized to ensure each dataset has the same total area, and the `Others' group is set to 5\% of the total size of the top genera.}
    \label{fig:treemapchartGenus}
\end{figure*}

\begin{figure*}[!ht]
    \centering
    \includegraphics[width=\textwidth]{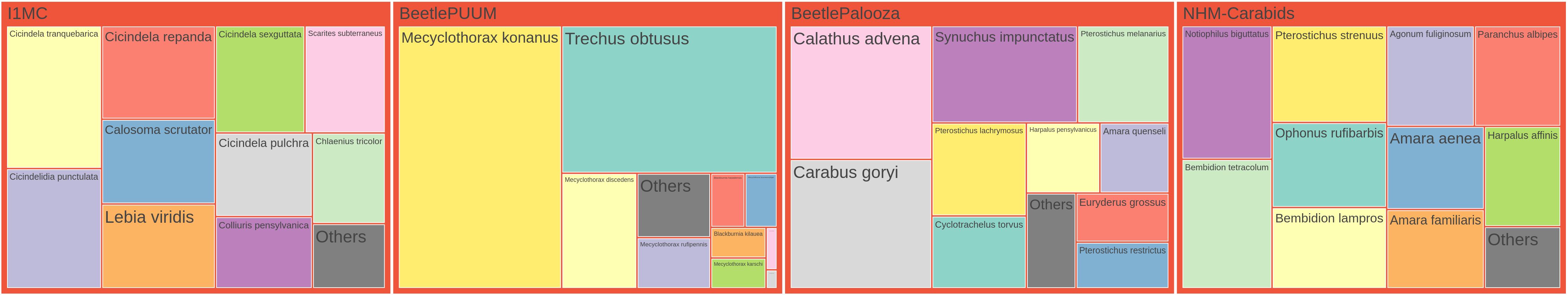}
    \captionsetup{skip=2pt}
    \caption{Treemap visualization of species distribution across four datasets, presented in the same format as Figure ~\ref{fig:treemapchartGenus}. For each dataset, the top 10 species by count are displayed individually, with all other species grouped into an `Others' category. The area of each rectangle is normalized to ensure equal total area per dataset, and the `Others' category is scaled to 5\% of the total size of the top species.}
    \label{fig:treemapchartSpecies}
\end{figure*}

\paragraph{Treemap Visualization.}Figures ~\ref{fig:treemapchartGenus} and ~\ref{fig:treemapchartSpecies} show the distribution of genera and species across four datasets through treemap visualizations. These hierarchical visualizations represent taxonomic abundance data where rectangle sizes correspond to the relative frequency of each taxon. In both figures, only the top 10 taxa are displayed individually for each dataset, with remaining taxa consolidated into an `Others' category. The visualizations are normalized to ensure comparable area allocation across datasets while maintaining the relative proportions within each dataset. This representation allows for immediate visual identification of dominant taxa in each dataset and facilitates cross-dataset comparison of taxonomic composition patterns.

\subsection{Pretrained Vision Encoders}
\label{app_subsec:model-details}
Our evaluation includes three model categories to provide comprehensive insights into representation learning for taxonomy. Vision-language models offer potential semantic alignment between visual features and taxonomic concepts through natural language grounding. Self-supervised models present the advantage of learning robust visual representations without requiring extensive labeled data, which is particularly valuable given the taxonomic annotation bottleneck. And lastly, vision-only supervised models serve as important baselines representing the conventional approach to visual classification tasks. By systematically comparing these complementary paradigms, we can identify which fundamental learning approaches best capture the hierarchical and fine-grained distinctions for classification.

\subsubsection{Data Preparation}
\label{app_subsec:data-prepare}
First, we filter and clean the dataset so that it contains only the images and corresponding genus and species labels, with missing labels designated as \textit{Unknown}. For feature extraction, images were processed using a pretrained vision model. Each image was loaded, converted to RGB, and passed through a transformation pipeline—where images were resized to $224 \times 224$ pixels, converted to tensors, and normalized using a mean of $[0.485, 0.456, 0.406]$ and standard deviation of $[0.229, 0.224, 0.225]$. These preprocessing parameters follow standard practices used for models trained on the ImageNet dataset. The transformed images were then fed into the model, and features were extracted from the last hidden state, averaged across the sequence dimension, producing a 768-dimensional feature vector per image. The dataset was divided into labeled and unlabeled samples. Features for labeled samples were extracted and stacked into a matrix, with labels encoded as integers using a label encoder. A train-test split was applied to the labeled samples, using a train-test ratio of 0.80 to 0.20. Features for unlabeled samples were similarly extracted and combined with the labeled test set to form the final test feature matrix. To prepare the features for modeling, standardization was performed using a standard scaler. The scaler was fitted on the training features to compute the mean and variance, then applied to both training and test features, ensuring zero mean and unit variance across all dimensions. This step optimizes the data for downstream machine learning algorithms sensitive to feature scaling.

\section{Result Details}
\label{app:results-details}

\subsection{Performance Evaluation Metrics}
\label{app_subsubsec:performance_metrics}
For our performance analysis, we select the Matthews Correlation Coefficient (MCC) as one of our primary evaluation metrics due to its robustness in handling significant class imbalance, a key characteristic of the datasets we use. For instance, some genera in our study comprise over 14,000 specimens, while others are represented by fewer than 5. Unlike simpler metrics, MCC provides a balanced assessment by integrating all elements of the confusion matrix—true positives (TP), true negatives (TN), false positives (FP), and false negatives (FN) — making it particularly well-suited for taxonomic identification tasks. In such tasks, accurately classifying rare taxa is just as critical as identifying common ones, and MCC’s sensitivity to all four components ensures a comprehensive evaluation \cite{boughorbel2017optimal, chicco2020advantages, chicco2021matthews}. For completeness, we also report the four baseline performance metrics- accuracy, precision, recall, and F1-score, to provide a broader perspective on model performance. Additionally, given the long-tailed nature of our datasets, we calculate macro-accuracy to better reflect performance across all classes. Macro-accuracy averages the accuracy for each class without weighting by class size, offering a clearer picture of the model’s ability to handle underrepresented taxa. This complements the MCC by emphasizing equitable performance across the dataset’s skewed distribution, ensuring that our evaluation captures both overall effectiveness and fairness in classification. 

\subsection{Benchmarking}
\label{app_subsubsec:benchmarking}
Our comprehensive evaluation of vision and vision language models reveals significant performance patterns. Tables~\ref{tab:benchmark_results} and~\ref{tab:benchmark_results_macro} present the micro and macro accuracy scores respectively, for genus and species classification across all datasets, while Table~\ref{tab:benchmark_details} provides a breakdown of multiple performance metrics. From the scores, we see that vision language models consistently outperform other approaches, with ViLT demonstrating superior performance across all datasets and metrics. ViLT achieves perfect genus-level accuracy (1.0) and exceptional species identification (0.997) on smaller, curated collections like BeetlePUUM, with corresponding perfect MCC scores (1.0 for genus, 0.995 for species). It maintains decent performance even on the challenging I1MC dataset (0.891 genus, 0.763 species micro-accuracy; MCC scores of 0.889 and 0.763 respectively), confirming that the integration of visual and textual features provides powerful taxonomic discrimination capabilities. Among other vision-language models, BioCLIP consistently ranks second, showing particularly strong performance on curated datasets but experiencing a more significant performance drop on larger, more heterogeneous collections. CLIP and SigLIP follow similar patterns but with lower performance scores.\\
Model's performance generally declines as the dataset size and heterogeneity increase, with all models showing a marked reduction in species-level identification accuracy on larger datasets. For instance, while ViLT maintains high precision and recall (both $>$0.99) for both genus and species on BeetlePUUM and BeetlePalooza, these metrics decline to approximately 0.74 for species classification on the I1MC dataset. Among vision-only models, supervised approaches (particularly BeIT and ConvNeXt) outperform self-supervised alternatives. BeIT achieves the best results in its category (0.923 genus, 0.821 species micro-accuracy on the merged dataset; MCC scores of 0.919 and 0.821), indicating that representations pretrained on general image collections transfer effectively to specialized taxonomic tasks. Within self-supervised models, DINOv2 consistently leads (with MCC scores reaching 0.968 for genus classification on BeetlePUUM), though it falls short of both vision-language models and supervised vision models. The performance gap between genus and species classification widens considerably in larger datasets, highlighting the increasing difficulty of fine-grained classification as taxonomic specificity increases. This pattern is consistent across all model types, with species-level F1 scores typically 10-30\% points lower than genus-level scores on the larger datasets. \\
Macro-accuracy scores reveal similar patterns, but all models show considerably lower macro-accuracy compared to micro-accuracy, particularly for species-level classification, highlighting significant class imbalance challenges in long-tailed datasets. This disparity is most pronounced in larger, more diverse datasets like I1MC, where even the top-performing ViLT model shows a substantial gap between micro-accuracy (0.763) and macro-accuracy (0.546) for species classification. The gap between genus and species classification is further amplified in macro-accuracy metrics. For example, on the merged dataset, ViLT achieves a genus macro-accuracy of 0.783 but drops to 0.657 for species classification, indicating that models struggle particularly with rare or underrepresented species. This pattern holds across all categories but is most severe for vision-only self-supervised models, where DINOv2's species macro-accuracy reaches only 0.391 on the merged dataset.

\subsection{Sample Efficient Probing}
\label{app_subsubsec:probing}
To assess sample efficiency and evaluate the cost-performance trade-offs for long-tailed datasets, we benchmark six leading vision models (ViLT, BioCLIP, ConvNeXt, CLIP, SWINv2, LeViT) across multiple dataset sizes and two sampling strategies: Balanced Sampling and Proportional Sampling. Balanced Sampling ensures equal representation across taxa, whereas Proportional Sampling maintains the natural class distribution, aligning with real-world imbalances. In our experimental design, we implemented both approaches across three dataset sizes. With Balanced Sampling, we extracted precisely 10, 20, and 50 images per species, resulting in total datasets of 2,900, 5,800, and 14,500 images respectively (across 290 species). For the corresponding Proportional Sampling datasets, we maintained identical total image counts (2,900, 5,800, and 14,500) but distributed them according to the natural frequency of each species in the source collection. This parallel sampling approach allowed us to evaluate classification performance under both artificial balance and natural distribution conditions, providing insight into model robustness across varying levels of class imbalance. The Balanced approach addresses potential bias against rare taxa, while the Proportional approach better reflects deployment conditions where certain species occur more frequently than others.
\begin{figure*}[!t]
\centering
\includegraphics[width=\textwidth]{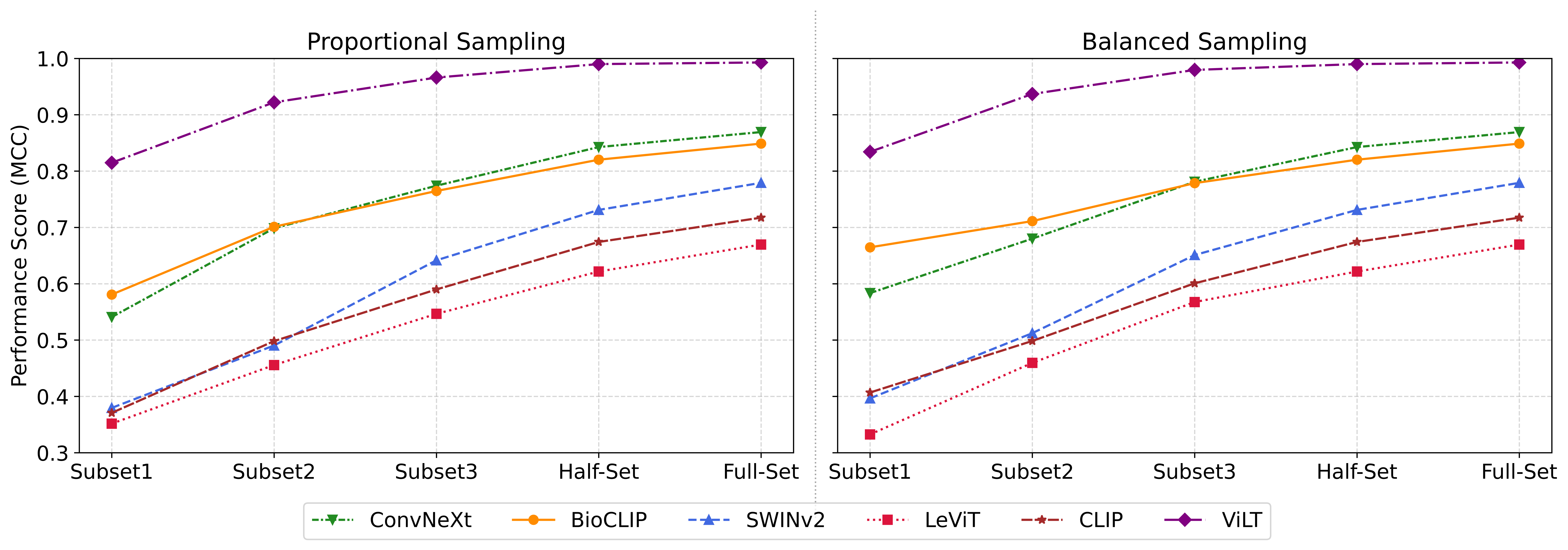}
\captionsetup{skip=3pt}
\caption{Performance of vision models under Proportional and Balanced Sampling strategies across increasing dataset sizes, highlighting sample efficiency and the impact of sampling on model performance. The models show a steep initial improvement with smaller subsets, followed by a plateau in performance as dataset size grows, indicating diminishing returns with scale. ViLT outperforms the rest.}
\label{fig:probing_trends}
\end{figure*}

As illustrated in Figure~\ref{fig:probing_trends} and Table~\ref{tab:probing_details}, ViLT consistently outperformed all other models across both sampling strategies and all dataset sizes, achieving near-ceiling performance with full supervision (Acc 0.9929, MCC 0.9928). Notably, even on small balanced subsets (e.g., Subset1 with 2900 images), ViLT achieved a strong accuracy of 0.8345, with a +0.107 jump in accuracy between the Subset1 (2900 images) and Subset2 (5800 images). However, the performance gains quickly diminished with scale, culminating in a marginal +0.003 improvement when scaling from the Half-set to the Full-set. This highlights an important insight: while adding more data improves results, the performance gains become progressively smaller, implying that strategically selected or subsampled training data - especially with balanced representation - can lead to competitive or even near-optimal performance without the computational burden of full-scale training. Furthermore, we observe that models exhibit varied sensitivity to sampling strategy. While ViLT maintained strong performance across both strategies, models like CLIP and LeViT performed notably worse under Proportional Sampling, suggesting that class imbalance exacerbates weaknesses in certain architectures. These findings provide a critical guide for practitioners working with long-tailed or resource-constrained settings: strategic subsampling can yield high-performance outcomes with significantly reduced data requirements, reinforcing the need for thoughtful dataset design over brute-force scaling.

\subsection{Cross-Dataset Domain Adaptation}
\label{app_subsubsec:domainshift}
The evaluation of pretrained vision models for cross-dataset domain adaptation reveals significant insights into their generalizability, particularly in the context of taxonomic classification across lab and in-situ imaging domains. Our experiments highlight the challenges and varying performance levels when adapting models between curated lab collections (NHM-Carabids and BeetlePalooza) and in-situ field images (I1MC). In the lab-to-lab adaptation scenario, where models were trained on NHM-Carabids and tested on BeetlePalooza at the genus level, ViLT demonstrated exceptional performance with an accuracy of 0.9230 and an MCC of 0.9106 across 16 shared taxa. This high performance underscores ViLT’s ability to generalize effectively between lab-based datasets, likely due to the controlled imaging conditions and taxonomic consistency between NHM and BeetlePalooza. Conversely, the lab-to-in-situ adaptation scenarios - training on NHM or BeetlePalooza and testing on I1MC - revealed a marked decline in performance across all models, reflecting the challenge of adapting from controlled lab settings to the variable conditions of field images. When trained on NHM and tested on I1MC at the genus level (57 taxa), ViLT again outperformed others with an accuracy of 0.6907 and an MCC of 0.6736, though these scores are notably lower than in the lab-to-lab case. At the species level (68 taxa), ViLT’s accuracy dropped to 0.5740 with an MCC of 0.5680, highlighting the increased difficulty of fine-grained classification in in-situ contexts. Training on BPZ and testing on I1MC produced similar trends. At the genus level (33 taxa), ViLT achieved an accuracy of 0.6001 and an MCC of 0.5756; at the species level (72 taxa), ViLT’s accuracy was 0.4757 with an MCC of 0.4676. These results underscore the inherent difficulty of adapting to in-situ data, which is challenging by nature due to uncontrolled conditions. The drop in performance is further exacerbated by I1MC-specific limitations, including inconsistent image quality, frequent partial views of specimens, varied perspectives (dorsal, lateral, ventral, anterior), and issues with focus and clarity.

\section{Multi-Modal Feature Integration}
\label{app_sec:multimodality}
Fine-grained visual recognition often relies on more than just visual cues \cite{miao2024new, lurig2021computer, hu2020fine, xu2023improving}. As two of our used datasets contain morphological traits and environmental data, we examine \textit{how effectively incorporating these additional modalities enhances taxonomic classification.} To investigate this, we conduct experiments using the BeetlePalooza dataset and a 1000-specimen subset, comparing image-only classification to approaches that combine visual features with morphological measurements (elytral dimensions) and environment metadata (geographic coordinates, elevation).

\subsection*{Experiment Results}
\label{app_subsec:results_multimodality}
We evaluate four models - BioCLIP, ConvNeXt, DINOv2, and ViLT - across three data configurations: image-only, image and morphological traits (image+traits), and image, traits, and environmental data (image+traits+env). Summary of model performance on two data sets with four models across all modality configurations is presented in Table~\ref{tab:multimodality_details}. Scores show that for the 1,000-specimen subset, vision-only models showed mixed responses to additional modalities. DINOv2’s accuracy was 0.7750 with just images, dropping to 0.7600 with traits and further to 0.7550 with traits and environmental data, suggesting extra modalities were not helpful. ConvNeXt started at 0.8150 with images alone, improved slightly to 0.8350 with traits, but fell to 0.8000 with environmental data added, indicating inconsistent benefits. In contrast, vision-language models behaved differently. BioCLIP’s accuracy rose steadily from 0.8150 (image-only) to 0.8300 (image+traits) and 0.8450 (image+traits+env), showing consistent gains. ViLT, however, achieved a strong 0.9350 with images alone but remained unchanged with traits (0.9350) and dropped substantially to 0.9050 with environmental data, suggesting additional modalities may disrupt its performance. \\
On the full dataset, trends shifted. Vision-only models benefited more from multi-modal inputs at scale. DINOv2’s accuracy increased from 0.9496 (image-only) to 0.9478 (image+traits) and 0.9513 (image+traits+env), while ConvNeXt improved from 0.9531 (image-only) to 0.9566 (image+traits) and 0.9649 (image+traits+env), indicating that additional modalities became helpful with more data. For vision-language models, BioCLIP again showed steady improvement, rising from 0.9373 (image-only) to 0.9417 (image+traits) and 0.9579 (image+traits+env), reinforcing its ability to leverage extra data. ViLT, starting near-perfect at 0.9982 (image-only), dropped marginally to 0.9956 with both traits and traits+env, suggesting limited or negative impact from additional modalities. These results reveal distinct patterns. DINOv2 and ConvNeXt struggle to benefit from extra modalities in the subset but improve at full scale, possibly due to better generalization with larger data. BioCLIP consistently gains from multi-modal inputs across both scales, highlighting its robustness. ViLT, however, shows no benefit in the subset where it suffers a substantial drop, and a marginal decline at scale, possibly indicating saturation or sensitivity to non-visual data. Given these inconsistencies, we cannot draw a firm conclusion on the effectiveness of multi-modal integration. Further experiments, varying dataset sizes, modalities, and model architectures, are needed to clarify these trends and determine optimal strategies for taxonomic classification.

\section{Feature Mapping}
\label{app_subsec:feature-mapping}
\begin{figure*}[!t]
    \centering
    \includegraphics[width=0.88\textwidth, keepaspectratio]{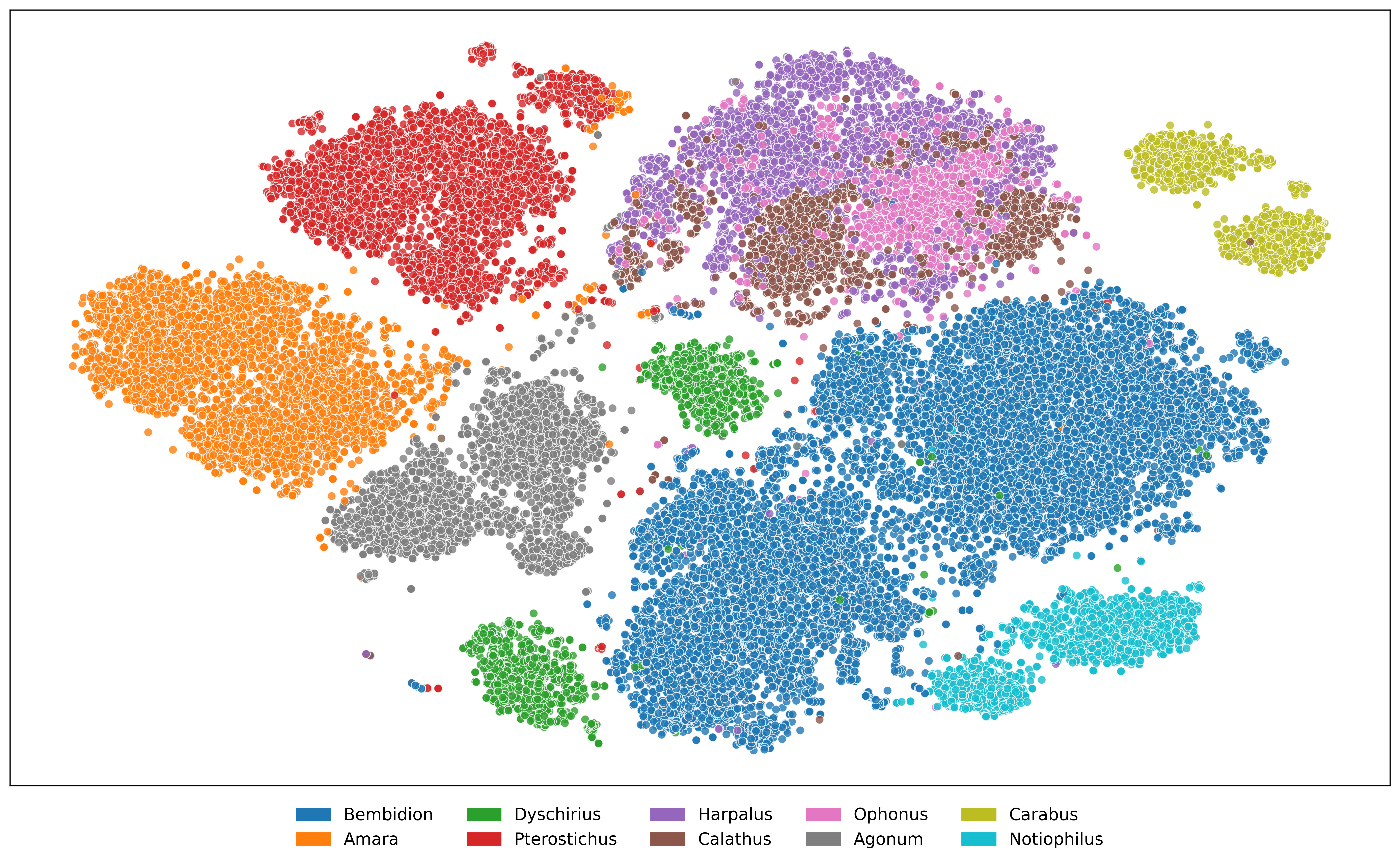}
    \captionsetup{skip=3pt}
    \caption{t-SNE visualization of feature embeddings extracted from ViLT for the top 10 genera in the NHM-Carabids dataset. Areas of overlap between genera suggest shared morphological traits that represent taxonomic challenges for automated identification systems.}
    \label{fig:genus_embeddings_nhm}
\end{figure*}

\begin{figure*}[!t]
    \centering
    \includegraphics[width=0.88\textwidth, keepaspectratio]{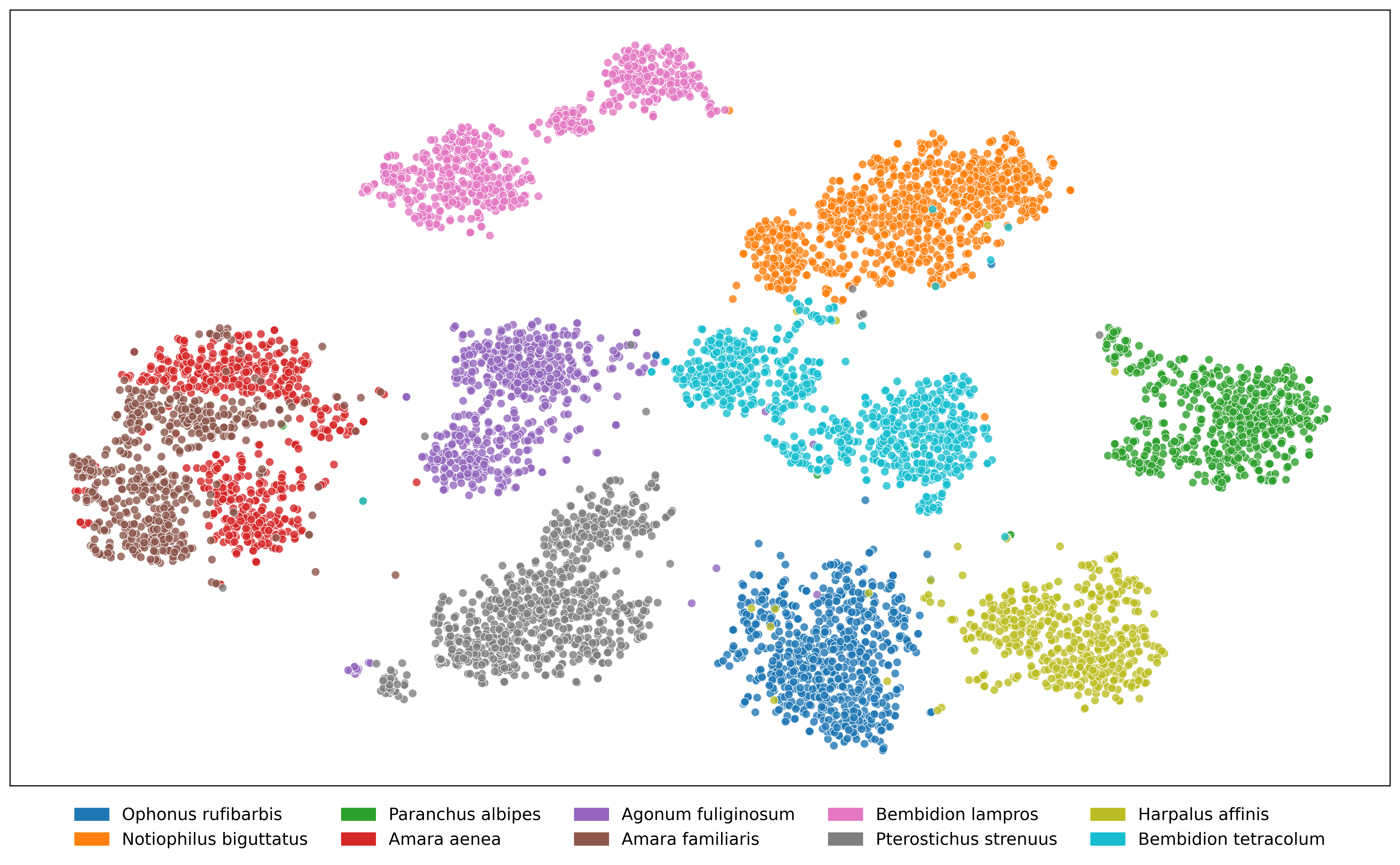}
    \captionsetup{skip=3pt}
    \caption{t-SNE visualization of feature embeddings extracted from ViLT for the top 10 species in the NHM-Carabids dataset. Some species (particularly within the same genus: Amara aenea and Amara familiaris) show partial overlap in feature space, indicating morphological similarities that challenge classification. The distinct separation between most clusters demonstrates the model's ability to capture species-specific visual characteristics despite intraspecific variation.}
    \label{fig:species_embeddings_nhm}
\end{figure*}

\begin{figure*}[!t]
    \centering
    \includegraphics[width=0.88\textwidth, keepaspectratio]{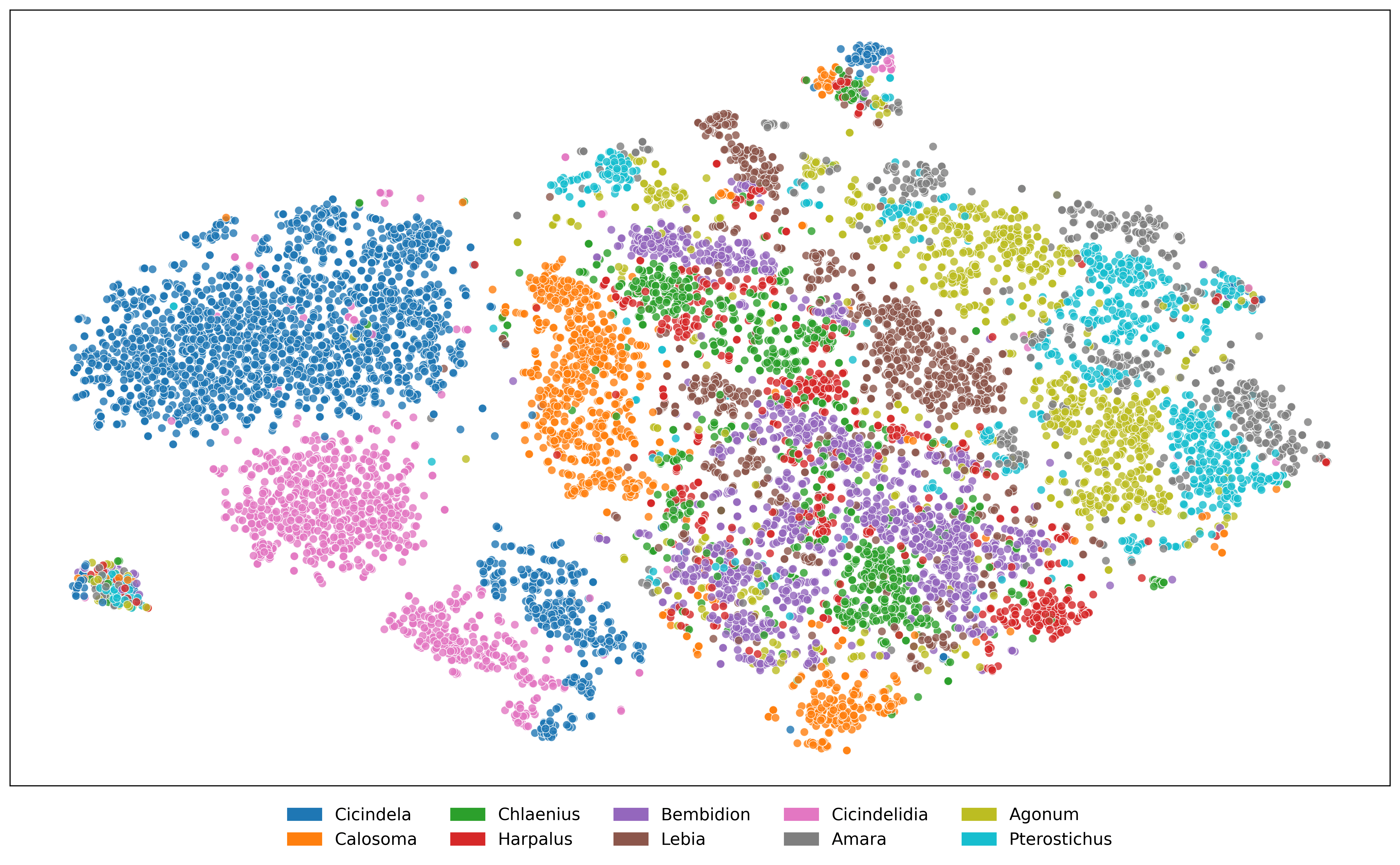}
    \captionsetup{skip=3pt}
    \caption{t-SNE visualization of feature embeddings extracted from ViLT for the top 10 genera in the I1MC dataset. Scattered and overlapping clusters imply that the model struggles to capture clear genus boundaries. High intra-genus variance and inter-genus proximity highlight the limitations of the embedding space, reflecting inconsistencies in data representation.}
    \label{fig:genus_embeddings_i1mc}
\end{figure*}

\begin{figure*}[!t]
    \centering
    \includegraphics[width=0.88\textwidth, keepaspectratio]{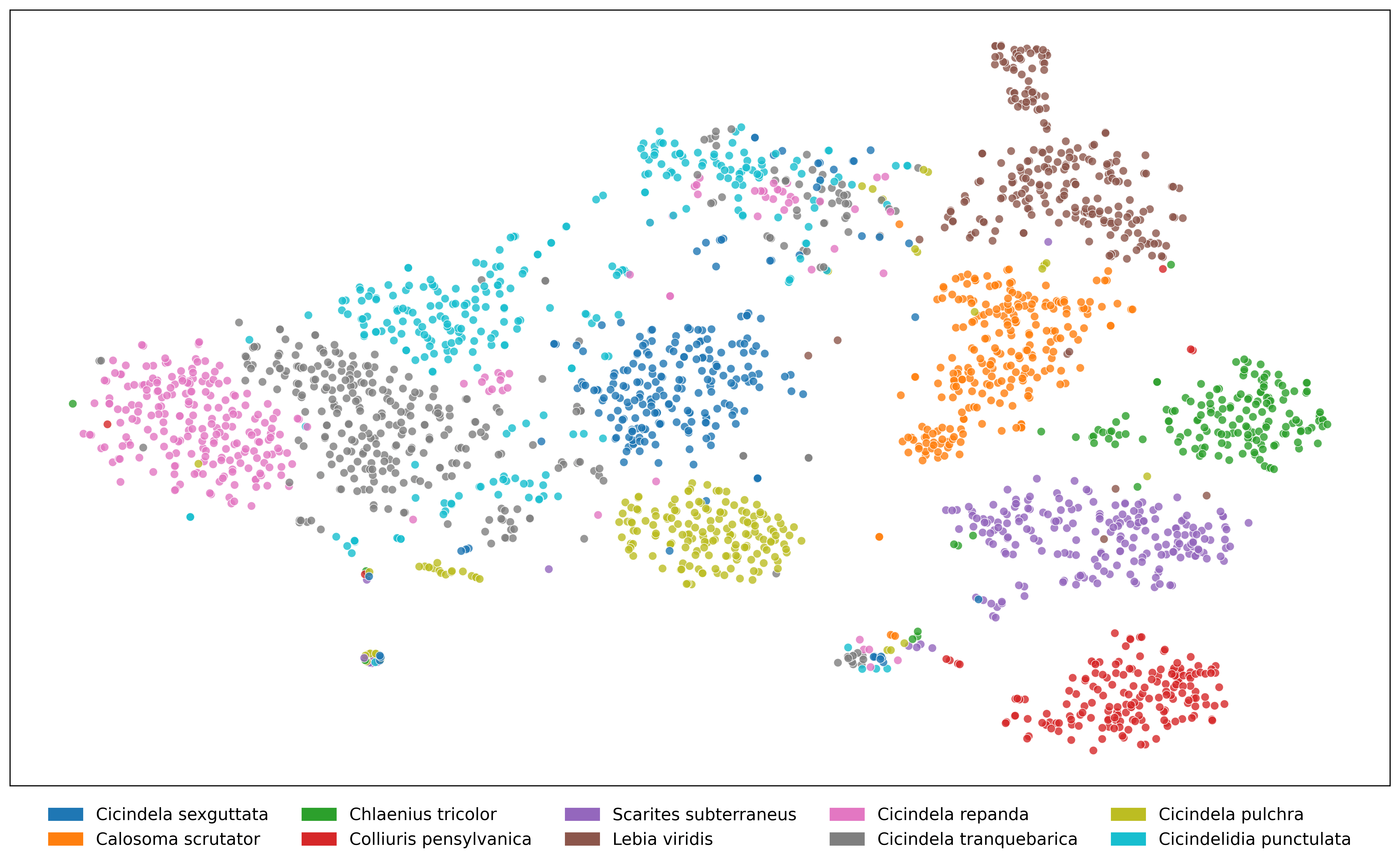}
    \captionsetup{skip=3pt}
    \caption{t-SNE visualization of feature embeddings for the top 10 species in the I1MC dataset, extracted using ViLT. Species within the genus Cicindela exhibit significant overlap, reflecting high morphological similarity within the genus. In contrast, species from other genera (e.g., Calosoma scrutator, Chlaenius tricolor) form well-separated clusters, indicating more distinctive visual features. This suggests that while the model captures genus-level distinctions well, species-level differentiation within certain genera remains a challenge.}
    \label{fig:species_embeddings_i1mc}
\end{figure*}

For better visualization of taxonomic relationships, we extract feature embeddings from pretrained vision models and apply dimensionality reduction techniques. These embeddings are derived from high-dimensional representations of the input data, capturing intricate patterns and characteristics that are not easily discernible in their raw form. To make these relationships more interpretable, we employ a dimensionality reduction method, t-SNE, that projects the high-dimensional embeddings into a two-dimensional space while preserving the underlying structure of the data as much as possible. The embeddings are then plotted to reveal distinct clustering patterns. In the plot, each cluster is represented by a unique color, with the legend indicating the corresponding genera, allowing for a clear visual interpretation of how closely related or distinct the groups are based on their feature representations. This mapping helps visualizing the effectiveness of pretrained models in capturing meaningful taxonomic differences among various taxa in a more intuitive manner. Such insights can guide further analysis, such as identifying potential misclassifications or discovering previously unrecognized similarities between genera. Figures~\ref{fig:genus_embeddings_nhm} and~\ref{fig:species_embeddings_nhm} illustrate how the pretrained model captures meaningful taxonomic structure, with clear cluster separation at both genus and species levels, and reveal cases of morphological similarity where overlap occurs in the embedding space. On the other hand, figures~\ref{fig:genus_embeddings_i1mc} and~\ref{fig:species_embeddings_i1mc} highlight the limitations of the I1MC dataset. In these visualizations, the model struggles to clearly separate genera and species, particularly at the genus level, where scattered and overlapping clusters suggest that the dataset's inherent variability makes it difficult for the model to capture distinct genus boundaries. This high intra-genus variance and inter-genus proximity emphasize the challenges of the dataset in providing clean and separable data representations. At the species level, overlap within genera Cicindela further underscores the dataset's complexity, as species within the same genus exhibit significant morphological similarity, making it harder for the model to differentiate them. From the accuracy scores in Tables~\ref{tab:benchmark_results} and~\ref{tab:benchmark_results_macro}, it is evident that the feature embeddings provide a prior signal of how performance is likely to unfold.

\clearpage

\clearpage
\clearpage

\begin{table*}[!t]
\centering
\resizebox{\textwidth}{!}{%
\setlength{\tabcolsep}{10pt}
\begin{tabular}{lcccccccccc}
\toprule
\multirow{2}{*}{\textbf{Model}} & \multicolumn{2}{c}{\textbf{BeetlePUUM}} & \multicolumn{2}{c}{\textbf{BeetlePalooza}} & \multicolumn{2}{c}{\textbf{NHM-London}} & \multicolumn{2}{c}{\textbf{I1MC}} & \multicolumn{2}{c}{\textbf{Merged-Dataset}} \\
\cmidrule(lr){2-3} \cmidrule(lr){4-5} \cmidrule(lr){6-7} \cmidrule(lr){8-9} \cmidrule(lr){10-11}
 & Genus & Species & Genus & Species & Genus & Species & Genus & Species & Genus & Species \\
\midrule
\midrule
\multicolumn{11}{l}{\textit{Vision-Language Models}} \\
\midrule
\underline{\textit{\textbf{ViLT}}} & \underline{\textbf{1.0000}} & \underline{\textbf{0.9969}} & \underline{\textbf{0.9987}} & \underline{\textbf{0.9982}} & \underline{\textbf{0.9984}} & \underline{\textbf{0.9950}} & \underline{\textbf{0.8905}} & \underline{\textbf{0.7633}} & \underline{\textbf{0.9715}} & \underline{\textbf{0.9397}} \\
BioCLIP & 0.9969 & 0.9292 & 0.9653 & 0.9376 & 0.9457 & 0.8498 & 0.7936 & 0.6095 & 0.9109 & 0.8054 \\
SigLIP & 0.9908 & 0.9323 & 0.9614 & 0.9328 & 0.9245 & 0.7864 & 0.6795 & 0.4852 & 0.8690 & 0.7372 \\
CLIP & 0.9815 & 0.8985 & 0.9310 & 0.8928 & 0.8725 & 0.7158 & 0.5483 & 0.3681 & 0.8037 & 0.6640 \\
\midrule
\midrule
\multicolumn{11}{l}{\textit{Vision-Only Self-Supervised Models}} \\
\midrule
\underline{\textit{\textbf{DINOv2}}} & \underline{0.9846} & 0.9108 & \underline{0.9715} & \underline{0.9499} & \underline{0.9367} & \underline{0.8106} & \underline{0.6440} & \underline{0.4426} & \underline{0.8663} & \underline{0.7352} \\
SwAV & 0.9846 & \underline{0.9231} & 0.9051 & 0.8660 & 0.8185 & 0.6582 & 0.4199 & 0.2571 & 0.7384 & 0.5928 \\
MoCov3 & 0.9723 & 0.8923 & 0.8853 & 0.8418 & 0.7355 & 0.5543 & 0.3967 & 0.2414 & 0.6727 & 0.5142 \\
ViTMAE & 0.9354 & 0.8369 & 0.8770 & 0.8336 & 0.7303 & 0.5387 & 0.3861 & 0.2152 & 0.6496 & 0.4762 \\
\midrule
\midrule
\multicolumn{11}{l}{\textit{Vision-Only Supervised Models}} \\
\midrule
\underline{\textit{\textbf{BeIT}}} & \underline{0.9969} & \underline{0.9354} & \underline{0.9798} & \underline{0.9592} & \underline{0.9673} & \underline{0.8876} & \underline{0.7641} & \underline{0.5720} & \underline{0.9225} & \underline{0.8213} \\
ConvNeXt & 0.9938 & 0.9385 & 0.9793 & 0.9534 & 0.9620 & 0.8785 & 0.7505 & 0.5409 & 0.9138 & 0.8060 \\
SWINv2 & 0.9692 & 0.8831 & 0.9618 & 0.9337 & 0.9105 & 0.7837 & 0.6425 & 0.4278 & 0.8511 & 0.7140 \\
LeViT & 0.9785 & 0.8985 & 0.9218 & 0.8779 & 0.8426 & 0.6719 & 0.5274 & 0.3306 & 0.7766 & 0.6171 \\
\bottomrule
\end{tabular}%
}
\captionsetup{skip=5pt}
\caption{Performance comparison of vision and vision-language models: Models are grouped by category, and ranked by \textbf{(micro)-accuracy} for genus and species classification. \textbf{\underline{Bold and Underlined}} values denote the highest score in each column across all models and \underline{Underlined} values refer to category-wise highest score in each column. \underline{\textit{Italicized and underlined}} text indicates the top model within each category; and \textbf{\underline{\textit{Bold, Italicized and underlined}}} text shows the best model across all categories.}
\label{tab:benchmark_results}
\end{table*} 
\begin{table*}[!t]
\centering
\resizebox{\textwidth}{!}{%
\setlength{\tabcolsep}{10pt}
\begin{tabular}{lcccccccccc}
\toprule
\multirow{2}{*}{\textbf{Model}} & \multicolumn{2}{c}{\textbf{BeetlePUUM}} & \multicolumn{2}{c}{\textbf{BeetlePalooza}} & \multicolumn{2}{c}{\textbf{NHM-London}} & \multicolumn{2}{c}{\textbf{I1MC}} & \multicolumn{2}{c}{\textbf{Merged-Dataset}} \\
\cmidrule(lr){2-3} \cmidrule(lr){4-5} \cmidrule(lr){6-7} \cmidrule(lr){8-9} \cmidrule(lr){10-11}
& Genus & Species & Genus & Species & Genus & Species & Genus & Species & Genus & Species \\
\midrule
\midrule
\multicolumn{11}{l}{\textit{Vision-Language Models}} \\
\midrule
\underline{\textit{\textbf{ViLT}}} & \underline{\textbf{1.0000}} & \underline{\textbf{0.9091}} & \underline{\textbf{0.9650}} & \underline{\textbf{0.9669}} & \underline{\textbf{0.9978}} & \underline{\textbf{0.9936}} & \underline{\textbf{0.6798}} & \underline{\textbf{0.5457}} & \underline{\textbf{0.7830}} & \underline{\textbf{0.6567}} \\
BioCLIP & 0.9984 & 0.6612 & 0.8434 & 0.7908 & 0.9219 & 0.8317 & 0.6037 & 0.4303 & 0.6669 & 0.4983 \\
SigLIP & 0.9936 & 0.6470 & 0.8875 & 0.7497 & 0.8940 & 0.7671 & 0.4451 & 0.3118 & 0.5968 & 0.4069 \\
CLIP & 0.9593 & 0.5406 & 0.8365 & 0.6764 & 0.8269 & 0.6906 & 0.3100 & 0.2106 & 0.5069 & 0.3281 \\
\midrule
\midrule
\multicolumn{11}{l}{\textit{Vision-Only Self-Supervised Models}} \\
\midrule
\underline{\textit{DINOv2}} & 0.9380 & \underline{0.5592} & \underline{0.9006} & \underline{0.7861} & \underline{0.9092} & \underline{0.7848} & \underline{0.4195} & \underline{0.2782} & \underline{0.5786} & \underline{0.3914} \\
SwAV & \underline{0.9625} & 0.5464 & 0.7616 & 0.6058 & 0.7679 & 0.6267 & 0.2406 & 0.1681 & 0.4135 & 0.2625 \\
MoCov3 & 0.9202 & 0.5060 & 0.7803 & 0.6130 & 0.6710 & 0.5173 & 0.2323 & 0.1415 & 0.3892 & 0.2339 \\
ViTMAE & 0.8485 & 0.3960 & 0.7797 & 0.5848 & 0.6384 & 0.5066 & 0.1893 & 0.1140 & 0.3490 & 0.1998 \\
\midrule
\midrule
\multicolumn{11}{l}{\textit{Vision-Only Supervised Models}} \\
\midrule
\underline{\textit{{BeIT}}} & \underline{{0.9984}} & \underline{{0.7980}} & \underline{{0.9189}} & \underline{{0.8082}} & \underline{{0.9533}} & \underline{{0.8744}} & \underline{{0.5686}} & \underline{{0.3899}} & \underline{{0.6979}} & \underline{{0.5007}} \\
ConvNeXt & 0.9936 & 0.7006 & 0.9075 & 0.7743 & 0.9464 & 0.8634 & 0.5359 & 0.3550 & 0.6880 & 0.4790 \\
SWINv2 & 0.9202 & 0.5451 & 0.8711 & 0.7633 & 0.8841 & 0.7576 & 0.4589 & 0.2712 & 0.5877 & 0.3890 \\
LeViT & 0.9300 & 0.4932 & 0.7757 & 0.6267 & 0.7997 & 0.6454 & 0.2887 & 0.1838 & 0.4855 & 0.2916 \\
\bottomrule
\end{tabular}%
}
\captionsetup{skip=5pt}
\caption{Performance comparison of vision and vision-language models: Models are grouped by category, and ranked by \textbf{macro-accuracy} for genus and species classification. \textbf{\underline{Bold and Underlined}} values denote the highest score in each column across all models and \underline{Underlined} values refer to category-wise highest score in each column. \underline{\textit{Italicized and underlined}} text indicates the top model within each category; and \textbf{\underline{\textit{Bold, Italicized and underlined}}} text shows the best model across all categories.}
\label{tab:benchmark_results_macro}
\end{table*}
\begin{table*}[!t]
\small
\centering
\setlength{\tabcolsep}{2pt}
\begin{adjustbox}{max width=\textwidth} 
\begin{tabular}{l*{10}{c}*{10}{c}}
\toprule
\toprule
\multirow{4.5}{*}{\textbf{Model}} & \multicolumn{10}{c}{\textbf{BeetlePUUM}} & \multicolumn{10}{c}{\textbf{BeetlePalooza}} \\
\cmidrule{2-21}
& \multicolumn{5}{c}{Genus} & \multicolumn{5}{c}{Species} & \multicolumn{5}{c}{Genus} & \multicolumn{5}{c}{Species} \\
\cmidrule{2-21}
& Acc & Pre & Rec & F1 & \multicolumn{1}{c}{MCC} & Acc & Pre & Rec & F1 & MCC & Acc & Pre & Rec & F1 & \multicolumn{1}{c}{MCC} & Acc & Pre & Rec & F1 & MCC \\
\midrule
\multicolumn{21}{l}{\textit{Vision-Language Models}} \\
\midrule
ViLT      & 1.0000 & 1.0000 & 1.0000 & 1.0000 & 1.0000 & 0.9969 & 0.9942 & 0.9969 & 0.9955 & 0.9954 & 0.9987 & 0.9983 & 0.9987 & 0.9985 & 0.9985 & 0.9982 & 0.9974 & 0.9982 & 0.9978 & 0.9981 \\
BioCLIP   & 0.9969 & 0.9972 & 0.9969 & 0.9970 & 0.9936 & 0.9292 & 0.9208 & 0.9292 & 0.9233 & 0.8917 & 0.9653 & 0.9651 & 0.9653 & 0.9633 & 0.9608 & 0.9376 & 0.9327 & 0.9376 & 0.9326 & 0.9333 \\
CLIP      & 0.9815 & 0.9816 & 0.9815 & 0.9815 & 0.9613 & 0.8985 & 0.8875 & 0.8985 & 0.8892 & 0.8435 & 0.9310 & 0.9312 & 0.9310 & 0.9302 & 0.9222 & 0.8928 & 0.8871 & 0.8928 & 0.8867 & 0.8854 \\
SigLIP    & 0.9908 & 0.9908 & 0.9908 & 0.9908 & 0.9808 & 0.9323 & 0.9280 & 0.9323 & 0.9281 & 0.8971 & 0.9614 & 0.9612 & 0.9614 & 0.9603 & 0.9564 & 0.9328 & 0.9232 & 0.9328 & 0.9256 & 0.9282 \\
\midrule
\multicolumn{21}{l}{\textit{Vision-Only Self-Supervised Models}} \\
\midrule
DINOv2    & 0.9846 & 0.9877 & 0.9846 & 0.9859 & 0.9678 & 0.9108 & 0.9060 & 0.9108 & 0.9064 & 0.8637 & 0.9715 & 0.9709 & 0.9715 & 0.9707 & 0.9678 & 0.9499 & 0.9456 & 0.9499 & 0.9449 & 0.9465 \\
ViTMAE    & 0.9354 & 0.9349 & 0.9354 & 0.9345 & 0.8636 & 0.8369 & 0.8212 & 0.8369 & 0.8228 & 0.7469 & 0.8770 & 0.8766 & 0.8770 & 0.8744 & 0.8612 & 0.8336 & 0.8176 & 0.8336 & 0.8206 & 0.8216 \\
SwAV      & 0.9846 & 0.9847 & 0.9846 & 0.9846 & 0.9678 & 0.9231 & 0.9162 & 0.9231 & 0.9140 & 0.8824 & 0.9051 & 0.9075 & 0.9051 & 0.9034 & 0.8931 & 0.8660 & 0.8539 & 0.8660 & 0.8519 & 0.8565 \\
MoCov3    & 0.9723 & 0.9723 & 0.9723 & 0.9720 & 0.9429 & 0.8923 & 0.8702 & 0.8923 & 0.8793 & 0.8349 & 0.8853 & 0.8868 & 0.8853 & 0.8831 & 0.8704 & 0.8418 & 0.8294 & 0.8418 & 0.8306 & 0.8305 \\
\midrule
\multicolumn{21}{l}{\textit{Vision-Only Supervised Models}} \\
\midrule
ConvNeXt  & 0.9938 & 0.9939 & 0.9938 & 0.9938 & 0.9871 & 0.9385 & 0.9329 & 0.9385 & 0.9350 & 0.9065 & 0.9793 & 0.9792 & 0.9793 & 0.9784 & 0.9767 & 0.9534 & 0.9428 & 0.9534 & 0.9464 & 0.9502 \\
SWINv2    & 0.9692 & 0.9695 & 0.9692 & 0.9687 & 0.9353 & 0.8831 & 0.8762 & 0.8831 & 0.8749 & 0.8198 & 0.9618 & 0.9611 & 0.9618 & 0.9608 & 0.9569 & 0.9337 & 0.9261 & 0.9337 & 0.9269 & 0.9291 \\
BeIT      & 0.9969 & 0.9970 & 0.9969 & 0.9969 & 0.9936 & 0.9354 & 0.9327 & 0.9354 & 0.9324 & 0.9024 & 0.9798 & 0.9790 & 0.9798 & 0.9791 & 0.9772 & 0.9592 & 0.9544 & 0.9592 & 0.9541 & 0.9564 \\
LeViT     & 0.9785 & 0.9780 & 0.9785 & 0.9780 & 0.9549 & 0.8985 & 0.8873 & 0.8985 & 0.8890 & 0.8434 & 0.9218 & 0.9231 & 0.9218 & 0.9203 & 0.9118 & 0.8779 & 0.8697 & 0.8779 & 0.8687 & 0.8694 \\
\bottomrule
\midrule
\multirow{4.5}{*}{\textbf{Model}} & \multicolumn{10}{c}{\textbf{NHM-Carabids}} & \multicolumn{10}{c}{\textbf{I1MC}} \\
\cmidrule{2-21}
& \multicolumn{5}{c}{Genus} & \multicolumn{5}{c}{Species} & \multicolumn{5}{c}{Genus} & \multicolumn{5}{c}{Species} \\
\cmidrule{2-21}
& Acc & Pre & Rec & F1 & \multicolumn{1}{c}{MCC} & Acc & Pre & Rec & F1 & MCC & Acc & Pre & Rec & F1 & \multicolumn{1}{c}{MCC} & Acc & Pre & Rec & F1 & MCC \\
\midrule
\multicolumn{21}{l}{\textit{Vision-Language Models}} \\
\midrule
ViLT      & 0.9984 & 0.9984 & 0.9984 & 0.9984 & 0.9983 & 0.9950 & 0.9951 & 0.9950 & 0.9950 & 0.9950 & 0.8905 & 0.8883 & 0.8905 & 0.8855 & 0.8867 & 0.7633 & 0.7420 & 0.7633 & 0.7370 & 0.7626 \\
BioCLIP   & 0.9457 & 0.9461 & 0.9457 & 0.9457 & 0.9417 & 0.8498 & 0.8533 & 0.8498 & 0.8498 & 0.8490 & 0.7936 & 0.7983 & 0.7936 & 0.7879 & 0.7865 & 0.6095 & 0.5977 & 0.6095 & 0.5800 & 0.6082 \\
CLIP      & 0.8725 & 0.8734 & 0.8725 & 0.8723 & 0.8630 & 0.7158 & 0.7221 & 0.7158 & 0.7158 & 0.7143 & 0.5483 & 0.5472 & 0.5483 & 0.5391 & 0.5321 & 0.3681 & 0.3663 & 0.3681 & 0.3442 & 0.3659 \\
SigLIP    & 0.9245 & 0.9248 & 0.9245 & 0.9243 & 0.9189 & 0.7864 & 0.7921 & 0.7864 & 0.7866 & 0.7852 & 0.6795 & 0.6784 & 0.6795 & 0.6714 & 0.6681 & 0.4852 & 0.4660 & 0.4852 & 0.4527 & 0.4835 \\
\midrule
\multicolumn{21}{l}{\textit{Vision-Only Self-Supervised Models}} \\
\midrule
DINOv2    & 0.9367 & 0.9372 & 0.9367 & 0.9366 & 0.9320 & 0.8106 & 0.8155 & 0.8106 & 0.8103 & 0.8096 & 0.6440 & 0.6388 & 0.6440 & 0.6353 & 0.6314 & 0.4426 & 0.4468 & 0.4426 & 0.4189 & 0.4407 \\
ViTMAE    & 0.7303 & 0.7308 & 0.7303 & 0.7299 & 0.7104 & 0.5387 & 0.5439 & 0.5387 & 0.5385 & 0.5363 & 0.3792 & 0.3805 & 0.3792 & 0.3679 & 0.3559 & 0.2108 & 0.2127 & 0.2108 & 0.1910 & 0.2076 \\
SwAV      & 0.8185 & 0.8192 & 0.8185 & 0.8184 & 0.8051 & 0.6582 & 0.6660 & 0.6582 & 0.6569 & 0.6564 & 0.4172 & 0.4144 & 0.4172 & 0.4041 & 0.3954 & 0.2571 & 0.2491 & 0.2571 & 0.2314 & 0.2542 \\
MoCov3    & 0.7355 & 0.7372 & 0.7355 & 0.7353 & 0.7162 & 0.5543 & 0.5590 & 0.5543 & 0.5526 & 0.5519 & 0.3967 & 0.3916 & 0.3967 & 0.3850 & 0.3740 & 0.2411 & 0.2407 & 0.2411 & 0.2231 & 0.2382 \\
\midrule
\multicolumn{21}{l}{\textit{Vision-Only Supervised Models}} \\
\midrule
ConvNeXt  & 0.9620 & 0.9623 & 0.9620 & 0.9620 & 0.9591 & 0.8785 & 0.8819 & 0.8785 & 0.8784 & 0.8779 & 0.7505 & 0.7530 & 0.7505 & 0.7443 & 0.7417 & 0.5409 & 0.5385 & 0.5409 & 0.5164 & 0.5395 \\
SWINv2    & 0.9105 & 0.9111 & 0.9105 & 0.9105 & 0.9039 & 0.7837 & 0.7902 & 0.7837 & 0.7845 & 0.7825 & 0.6425 & 0.6521 & 0.6425 & 0.6394 & 0.6299 & 0.4278 & 0.4437 & 0.4278 & 0.4103 & 0.4259 \\
BeIT      & 0.9673 & 0.9675 & 0.9673 & 0.9673 & 0.9649 & 0.8876 & 0.8899 & 0.8876 & 0.8874 & 0.8870 & 0.7641 & 0.7661 & 0.7641 & 0.7582 & 0.7559 & 0.5720 & 0.5581 & 0.5720 & 0.5400 & 0.5706 \\
LeViT     & 0.8426 & 0.8440 & 0.8426 & 0.8423 & 0.8309 & 0.6719 & 0.6784 & 0.6719 & 0.6716 & 0.6702 & 0.5274 & 0.5270 & 0.5274 & 0.5174 & 0.5104 & 0.3306 & 0.3159 & 0.3306 & 0.3022 & 0.3283 \\
\bottomrule
\midrule
\multirow{4.5}{*}{\textbf{Model}} & \multicolumn{10}{c}{\textbf{I1MCv2}} & \multicolumn{10}{c}{\textbf{Merged-Dataset}} \\
\cmidrule{2-21}
& \multicolumn{5}{c}{Genus} & \multicolumn{5}{c}{Species} & \multicolumn{5}{c}{Genus} & \multicolumn{5}{c}{Species} \\
\cmidrule{2-21}
& Acc & Pre & Rec & F1 & \multicolumn{1}{c}{MCC} & Acc & Pre & Rec & F1 & MCC & Acc & Pre & Rec & F1 & \multicolumn{1}{c}{MCC} & Acc & Pre & Rec & F1 & MCC \\
\midrule
\multicolumn{21}{l}{\textit{Vision-Language Models}} \\
\midrule
ViLT      & 0.8928 & 0.8907 & 0.8928 & 0.8879 & 0.8890 & 0.7638 & 0.7431 & 0.7638 & 0.7379 & 0.7631 & 0.9715 & 0.9709 & 0.9715 & 0.9706 & 0.9701 & 0.9397 & 0.9383 & 0.9397 & 0.9350 & 0.9394 \\
BioCLIP   & 0.7936 & 0.7983 & 0.7936 & 0.7879 & 0.7865 & 0.6095 & 0.5977 & 0.6095 & 0.5800 & 0.6082 & 0.9109 & 0.9103 & 0.9109 & 0.9089 & 0.9067 & 0.8054 & 0.7993 & 0.8054 & 0.7958 & 0.8047 \\
CLIP      & 0.5483 & 0.5472 & 0.5483 & 0.5391 & 0.5321 & 0.3681 & 0.3663 & 0.3681 & 0.3442 & 0.3659 & 0.8037 & 0.8011 & 0.8037 & 0.7997 & 0.7944 & 0.6640 & 0.6573 & 0.6640 & 0.6529 & 0.6627 \\
SigLIP    & 0.6812 & 0.6802 & 0.6812 & 0.6736 & 0.6699 & 0.4862 & 0.4686 & 0.4862 & 0.4543 & 0.4845 & 0.8690 & 0.8677 & 0.8690 & 0.8670 & 0.8628 & 0.7372 & 0.7335 & 0.7372 & 0.7281 & 0.7362 \\
\midrule
\multicolumn{21}{l}{\textit{Vision-Only Self-Supervised Models}} \\
\midrule
DINOv2    & 0.6418 & 0.6394 & 0.6418 & 0.6336 & 0.6291 & 0.4426 & 0.4494 & 0.4426 & 0.4195 & 0.4407 & 0.8663 & 0.8658 & 0.8663 & 0.8641 & 0.8599 & 0.7352 & 0.7345 & 0.7352 & 0.7274 & 0.7342 \\
ViTMAE    & 0.3861 & 0.3768 & 0.3861 & 0.3700 & 0.3625 & 0.2152 & 0.2138 & 0.2152 & 0.1943 & 0.2121 & 0.6496 & 0.6499 & 0.6496 & 0.6481 & 0.6332 & 0.4762 & 0.4763 & 0.4762 & 0.4697 & 0.4741 \\
SwAV      & 0.4199 & 0.4176 & 0.4199 & 0.4081 & 0.3983 & 0.2554 & 0.2520 & 0.2554 & 0.2329 & 0.2526 & 0.7384 & 0.7368 & 0.7384 & 0.7353 & 0.7259 & 0.5928 & 0.5910 & 0.5928 & 0.5843 & 0.5913 \\
MoCov3    & 0.3962 & 0.3925 & 0.3962 & 0.3854 & 0.3736 & 0.2414 & 0.2391 & 0.2414 & 0.2232 & 0.2385 & 0.6727 & 0.6693 & 0.6727 & 0.6687 & 0.6575 & 0.5142 & 0.5059 & 0.5142 & 0.5036 & 0.5123 \\
\midrule
\multicolumn{21}{l}{\textit{Vision-Only Supervised Models}} \\
\midrule
ConvNeXt  & 0.7505 & 0.7530 & 0.7505 & 0.7443 & 0.7417 & 0.5409 & 0.5385 & 0.5409 & 0.5164 & 0.5395 & 0.9138 & 0.9137 & 0.9138 & 0.9124 & 0.9098 & 0.8060 & 0.8074 & 0.8060 & 0.8004 & 0.8053 \\
SWINv2    & 0.6425 & 0.6521 & 0.6425 & 0.6394 & 0.6299 & 0.4278 & 0.4437 & 0.4278 & 0.4103 & 0.4259 & 0.8511 & 0.8504 & 0.8511 & 0.8492 & 0.8441 & 0.7140 & 0.7117 & 0.7140 & 0.7063 & 0.7129 \\
BeIT      & 0.7641 & 0.7661 & 0.7641 & 0.7582 & 0.7559 & 0.5720 & 0.5581 & 0.5720 & 0.5400 & 0.5706 & 0.9225 & 0.9221 & 0.9225 & 0.9211 & 0.9189 & 0.8213 & 0.8202 & 0.8213 & 0.8147 & 0.8206 \\
LeViT     & 0.5274 & 0.5270 & 0.5274 & 0.5174 & 0.5104 & 0.3306 & 0.3159 & 0.3306 & 0.3022 & 0.3283 & 0.7766 & 0.7740 & 0.7766 & 0.7725 & 0.7661 & 0.6171 & 0.6086 & 0.6171 & 0.6048 & 0.6156 \\
\bottomrule
\end{tabular}
\end{adjustbox}
\captionsetup{skip=5pt}
\caption{Taxonomic Prediction at Genus and Species Level by Vision Models across all carabids datasets. Performance metrics include Accuracy (Acc), Precision (Pre), Recall (Rec), F1 score (F1), and Matthews Correlation Coefficient (MCC). NB. \textbf{I1MC-v2} is a version of the I1MC dataset where we kept the images NOT identified to genus/species level in the test set for future work}
\label{tab:benchmark_details}
\end{table*}
\begin{table*}[!t]
\small
\centering
\setlength{\tabcolsep}{5pt}
\begin{adjustbox}{max width=\textwidth}
\begin{tabular}{l*{14}{c}c}
\toprule
\multirow{3}{*}{Model} & \multicolumn{5}{c}{Subset1 (size: 2900)} & 
\multicolumn{5}{c}{Subset2 (size: 5800)} & 
\multicolumn{5}{c}{Subset3 (Size: 14500)} \\
\cmidrule(lr){2-6}\cmidrule(lr){7-11}\cmidrule(lr){12-16}
& Acc & Prec & Rec & F1 & MCC & 
Acc & Prec & Rec & F1 & MCC & 
Acc & Prec & Rec & F1 & MCC \\
\midrule
\midrule
\multicolumn{16}{c}{Balanced Sampling} \\
\midrule
ViLT     & 0.8345 & 0.8479 & 0.8345 & 0.8188 & 0.8341 & 
0.9371 & 0.9457 & 0.9371 & 0.9344 & 0.9369 & 
0.9797 & 0.9814 & 0.9797 & 0.9796 & 0.9796 \\
BioCLIP  & 0.6655 & 0.6826 & 0.6655 & 0.6455 & 0.6647 & 
0.7121 & 0.7415 & 0.7121 & 0.7071 & 0.7112 & 
0.7793 & 0.7936 & 0.7793 & 0.7784 & 0.7786 \\
ConvNeXt & 0.5845 & 0.5836 & 0.5845 & 0.5585 & 0.5834 & 
0.6810 & 0.7098 & 0.6810 & 0.6734 & 0.6801 & 
0.7817 & 0.7922 & 0.7817 & 0.7789 & 0.7810 \\
CLIP     & 0.4086 & 0.4093 & 0.4086 & 0.3839 & 0.4069 & 
0.5000 & 0.5351 & 0.5000 & 0.4924 & 0.4984 & 
0.6021 & 0.6160 & 0.6021 & 0.6005 & 0.6008 \\
SWINv2   & 0.3983 & 0.3921 & 0.3983 & 0.3712 & 0.3965 & 
0.5138 & 0.5516 & 0.5138 & 0.5081 & 0.5123 & 
0.6521 & 0.6680 & 0.6521 & 0.6502 & 0.6509 \\
LeViT    & 0.3345 & 0.3195 & 0.3345 & 0.3079 & 0.3325 & 
0.4612 & 0.4998 & 0.4612 & 0.4518 & 0.4595 & 
0.5690 & 0.5848 & 0.5690 & 0.5682 & 0.5675 \\
\midrule
\midrule
\multicolumn{16}{c}{Proportional Sampling} \\
\midrule
ViLT     & 0.8155 & 0.7740 & 0.8155 & 0.7782 & 0.8148 & 
0.9224 & 0.9211 & 0.9224 & 0.9144 & 0.9221 & 
0.9662 & 0.9676 & 0.9662 & 0.9646 & 0.9660 \\
BioCLIP  & 0.5828 & 0.5336 & 0.5828 & 0.5351 & 0.5808 & 
0.7026 & 0.7102 & 0.7026 & 0.6876 & 0.7012 & 
0.7659 & 0.7713 & 0.7659 & 0.7592 & 0.7647 \\
ConvNeXt & 0.5431 & 0.5047 & 0.5431 & 0.4994 & 0.5409 & 
0.7000 & 0.6968 & 0.7000 & 0.6809 & 0.6985 & 
0.7752 & 0.7842 & 0.7752 & 0.7704 & 0.7740 \\
CLIP     & 0.3741 & 0.3077 & 0.3741 & 0.3228 & 0.3708 & 
0.5009 & 0.4983 & 0.5009 & 0.4784 & 0.4982 & 
0.5921 & 0.5931 & 0.5921 & 0.5805 & 0.5899 \\
SWINv2   & 0.3828 & 0.3358 & 0.3828 & 0.3445 & 0.3796 & 
0.4931 & 0.4904 & 0.4931 & 0.4740 & 0.4904 & 
0.6434 & 0.6455 & 0.6434 & 0.6347 & 0.6416 \\
LeViT    & 0.3552 & 0.3121 & 0.3552 & 0.3151 & 0.3516 & 
0.4586 & 0.4524 & 0.4586 & 0.4385 & 0.4557 & 
0.5490 & 0.5600 & 0.5490 & 0.5418 & 0.5465 \\
\midrule
\midrule
\multirow{2}{*}{Model}  & \multicolumn{5}{c}{Half-set (Size: 30000)} & \multicolumn{5}{c}{Full-set (Size: 63077)} \\
\cmidrule(lr){2-6}\cmidrule(lr){7-11}
& Acc & Prec & Rec & F1 & MCC & Acc & Prec & Rec & F1 & MCC \\
\cmidrule(lr){2-6}\cmidrule(lr){7-11}
ViLT     & 0.9900 & 0.9905 & 0.9900 & 0.9900 & 0.9900 & 0.9929 & 0.9930 & 0.9929 & 0.9928 & 0.9928 \\
BioCLIP  & 0.8208 & 0.8274 & 0.8208 & 0.8207 & 0.8202 & 0.8496 & 0.8524 & 0.8496 & 0.8488 & 0.8488 \\
ConvNeXt & 0.8432 & 0.8491 & 0.8432 & 0.8434 & 0.8426 & 0.8699 & 0.8721 & 0.8699 & 0.8694 & 0.8693 \\
CLIP     & 0.6753 & 0.6847 & 0.6753 & 0.6757 & 0.6742 & 0.7188 & 0.7215 & 0.7188 & 0.7177 & 0.7173 \\
SWINv2   & 0.7320 & 0.7422 & 0.7320 & 0.7327 & 0.7311 & 0.7803 & 0.7832 & 0.7803 & 0.7792 & 0.7791 \\
LeViT    & 0.6230 & 0.6375 & 0.6230 & 0.6249 & 0.6217 & 0.6713 & 0.6749 & 0.6713 & 0.6700 & 0.6696 \\
\bottomrule
\end{tabular}
\end{adjustbox}
\captionsetup{skip=3pt}
\caption{Performance of vision models across Balanced Sampling (equal class representation) and Proportional Sampling (natural class distribution). Strategies and Varying Dataset Sizes (Subset1: 2900, Subset2: 5800, Subset3: 14500, Half-set: 30000, Full-set: 63077). Metrics Include Accuracy (Acc), Precision (Prec), Recall (Rec), F1-Score (F1), and Matthews Correlation Coefficient (MCC)}
\label{tab:probing_details}
\end{table*}
\begin{table*}[!t]
\small
\centering
\setlength{\tabcolsep}{8pt}
\begin{adjustbox}{max width=\textwidth}
\begin{tabular}{l*{9}{c}c}
\toprule
Case & Train & Test & Type & \#Taxa & Model & Accuracy & Precision & Recall & F1-Score & MCC \\
\midrule
\midrule
NHM-I1M-genus & NHM & I1M & genus & 57 & BioCLIP & 0.3899 & 0.6190 & 0.3899 & 0.4103 & 0.3623 \\
NHM-I1M-genus & NHM & I1M & genus & 57 & CLIP & 0.1946 & 0.3595 & 0.1946 & 0.1844 & 0.1425 \\
NHM-I1M-genus & NHM & I1M & genus & 57 & ConvNeXt & 0.3603 & 0.5438 & 0.3603 & 0.3610 & 0.3205 \\
NHM-I1M-genus & NHM & I1M & genus & 57 & LeViT & 0.2497 & 0.3919 & 0.2497 & 0.2510 & 0.2070 \\
NHM-I1M-genus & NHM & I1M & genus & 57 & SWINv2 & 0.3238 & 0.4822 & 0.3238 & 0.3165 & 0.2898 \\
NHM-I1M-genus & NHM & I1M & genus & 57 & ViLT & 0.6907 & 0.8168 & 0.6907 & 0.6966 & 0.6736 \\
\addlinespace
\hdashline
\addlinespace
NHM-I1M-species & NHM & I1M & species & 68 & BioCLIP & 0.4221 & 0.6546 & 0.4221 & 0.4362 & 0.4133 \\
NHM-I1M-species & NHM & I1M & species & 68 & CLIP & 0.0875 & 0.2436 & 0.0875 & 0.0856 & 0.0761 \\
NHM-I1M-species & NHM & I1M & species & 68 & ConvNeXt & 0.2291 & 0.4589 & 0.2291 & 0.2507 & 0.2204 \\
NHM-I1M-species & NHM & I1M & species & 68 & LeViT & 0.1120 & 0.1942 & 0.1120 & 0.1195 & 0.1026 \\
NHM-I1M-species & NHM & I1M & species & 68 & SWINv2 & 0.1750 & 0.3112 & 0.1750 & 0.1624 & 0.1618 \\
NHM-I1M-species & NHM & I1M & species & 68 & ViLT & 0.5740 & 0.7737 & 0.5740 & 0.6132 & 0.5680 \\
\addlinespace
\hdashline
\addlinespace
BPZ-I1M-genus & BPZ & I1M & genus & 33 & BioCLIP & 0.3553 & 0.5198 & 0.3553 & 0.3558 & 0.3257 \\
BPZ-I1M-genus & BPZ & I1M & genus & 33 & CLIP & 0.1386 & 0.2623 & 0.1386 & 0.1489 & 0.1071 \\
BPZ-I1M-genus & BPZ & I1M & genus & 33 & ConvNeXt & 0.3464 & 0.4843 & 0.3464 & 0.3430 & 0.3142 \\
BPZ-I1M-genus & BPZ & I1M & genus & 33 & LeViT & 0.1985 & 0.3115 & 0.1985 & 0.2042 & 0.1725 \\
BPZ-I1M-genus & BPZ & I1M & genus & 33 & SWINv2 & 0.3395 & 0.3881 & 0.3395 & 0.3202 & 0.2986 \\
BPZ-I1M-genus & BPZ & I1M & genus & 33 & ViLT & 0.6001 & 0.6931 & 0.6001 & 0.5823 & 0.5756 \\
\addlinespace
\hdashline
\addlinespace
BPZ-I1M-species & BPZ & I1M & species & 72 & BioCLIP & 0.3656 & 0.4298 & 0.3656 & 0.3400 & 0.3558 \\
BPZ-I1M-species & BPZ & I1M & species & 72 & CLIP & 0.1128 & 0.2835 & 0.1128 & 0.1107 & 0.1025 \\
BPZ-I1M-species & BPZ & I1M & species & 72 & ConvNeXt & 0.2592 & 0.3895 & 0.2592 & 0.2394 & 0.2498 \\
BPZ-I1M-species & BPZ & I1M & species & 72 & LeViT & 0.1422 & 0.1897 & 0.1422 & 0.1413 & 0.1276 \\
BPZ-I1M-species & BPZ & I1M & species & 72 & SWINv2 & 0.1913 & 0.3177 & 0.1913 & 0.1855 & 0.1802 \\
BPZ-I1M-species & BPZ & I1M & species & 72 & ViLT & 0.4757 & 0.4998 & 0.4757 & 0.4287 & 0.4676 \\
\addlinespace
\hdashline
\addlinespace
NHM-BPZ-genus & NHM & BPZ & genus & 16 & BioCLIP & 0.4632 & 0.7094 & 0.4632 & 0.5178 & 0.4222 \\
NHM-BPZ-genus & NHM & BPZ & genus & 16 & CLIP & 0.3076 & 0.4796 & 0.3076 & 0.2993 & 0.2161 \\
NHM-BPZ-genus & NHM & BPZ & genus & 16 & ConvNeXt & 0.3697 & 0.6106 & 0.3697 & 0.3770 & 0.3010 \\
NHM-BPZ-genus & NHM & BPZ & genus & 16 & LeViT & 0.3481 & 0.4962 & 0.3481 & 0.3284 & 0.2591 \\
NHM-BPZ-genus & NHM & BPZ & genus & 16 & SWINv2 & 0.4371 & 0.5217 & 0.4371 & 0.3886 & 0.3546 \\
NHM-BPZ-genus & NHM & BPZ & genus & 16 & \textbf{ViLT} & \textbf{0.9230} & \textbf{0.9552} & \textbf{0.9230} & \textbf{0.9311} & \textbf{0.9106} \\
\bottomrule
\end{tabular}
\end{adjustbox}
\captionsetup{skip=5pt}
\caption{Evaluation of Pretrained Vision Models for Cross-Dataset Domain Adaptation in Taxonomic Classification. This table reports performance metrics, including accuracy and Matthews Correlation Coefficient (MCC), alongside Accuracy (Acc), Precision (Prec), Recall (Rec), F1 Score (F1) - for models assessed in two domain adaptation scenarios: (1) lab-to-lab (NHM-Carabids to BeetlePalooza) and (2) lab-to-in-situ (NHM-Carabids or BeetlePalooza to I1MC). Results are presented at genus and species levels for taxa shared across source and target datasets, illustrating model generalizability across lab and field imaging contexts.}
\label{tab:domain_shift}
\end{table*}
\begin{table*}[!t]
\centering
\setlength{\tabcolsep}{8pt}
\begin{adjustbox}{max width=\textwidth}
\begin{tabular}{l*{8}{c}c}
\toprule
Dataset & Images & Data Type & Model & Acc & Prec & Rec & F1 & MCC \\
\midrule
\midrule
Subset & 1000 & image & BioCLIP & 0.8150 & 0.7585 & 0.8150 & 0.7780 & 0.8038 \\
Subset & 1000 & image & ConvNeXt & 0.8150 & 0.7733 & 0.8150 & 0.7864 & 0.8037 \\
Subset & 1000 & image & DINOv2 & 0.7750 & 0.7313 & 0.7750 & 0.7349 & 0.7609 \\
Subset & 1000 & image & ViLT & 0.9350 & 0.9121 & 0.9350 & 0.9172 & 0.9314 \\
\addlinespace
\hdashline
\addlinespace
Subset & 1000 & image+traits & BioCLIP & 0.8300 & 0.7603 & 0.8300 & 0.7850 & 0.8198 \\
Subset & 1000 & image+traits & ConvNeXt & 0.8350 & 0.7899 & 0.8350 & 0.8052 & 0.8251 \\
Subset & 1000 & image+traits & DINOv2 & 0.7600 & 0.7312 & 0.7600 & 0.7290 & 0.7449 \\
Subset & 1000 & image+traits & ViLT & 0.9350 & 0.9121 & 0.9350 & 0.9172 & 0.9314 \\
\addlinespace
\hdashline
\addlinespace
Subset & 1000 & image+traits+env & BioCLIP & 0.8450 & 0.7958 & 0.8450 & 0.8064 & 0.8347 \\
Subset & 1000 & image+traits+env & ConvNeXt & 0.8000 & 0.7304 & 0.8000 & 0.7528 & 0.7863 \\
Subset & 1000 & image+traits+env & DINOv2 & 0.7550 & 0.6896 & 0.7550 & 0.7030 & 0.7379 \\
Subset & 1000 & image+traits+env & ViLT & 0.9050 & 0.8783 & 0.9050 & 0.8814 & 0.8990 \\
\midrule
Full & 11372 & image & BioCLIP & 0.9373 & 0.9325 & 0.9373 & 0.9323 & 0.9330 \\
Full & 11372 & image & ConvNeXt & 0.9531 & 0.9426 & 0.9531 & 0.9461 & 0.9498 \\
Full & 11372 & image & DINOv2 & 0.9496 & 0.9453 & 0.9496 & 0.9446 & 0.9461 \\
Full & 11372 & image & ViLT & 0.9982 & 0.9974 & 0.9982 & 0.9978 & 0.9981 \\
\addlinespace
\hdashline
\addlinespace
Full & 11372 & image+traits & BioCLIP & 0.9417 & 0.9357 & 0.9417 & 0.9368 & 0.9375 \\
Full & 11372 & image+traits & ConvNeXt & 0.9566 & 0.9512 & 0.9566 & 0.9514 & 0.9535 \\
Full & 11372 & image+traits & DINOv2 & 0.9478 & 0.9445 & 0.9478 & 0.9430 & 0.9441 \\
Full & 11372 & image+traits & ViLT & 0.9956 & 0.9951 & 0.9956 & 0.9948 & 0.9953 \\
\addlinespace
\hdashline
\addlinespace
Full & 11372 & image+traits+env & BioCLIP & 0.9579 & 0.9531 & 0.9579 & 0.9536 & 0.9549 \\
Full & 11372 & image+traits+env & ConvNeXt & 0.9649 & 0.9604 & 0.9649 & 0.9604 & 0.9624 \\
Full & 11372 & image+traits+env & DINOv2 & 0.9513 & 0.9502 & 0.9513 & 0.9468 & 0.9479 \\
Full & 11372 & image+traits+env & ViLT & 0.9956 & 0.9952 & 0.9956 & 0.9950 & 0.9953 \\
\bottomrule
\end{tabular}
\end{adjustbox}
\captionsetup{skip=5pt}
\caption{Performance comparison of four models (BioCLIP, ConvNeXt, DINOv2, ViLT) on species-level classification using the BeetlePalooza dataset. Results are reported for both the full dataset and a 1,000-specimen subset across three input configurations: image-only, image with morphological traits (image+traits), and image with both traits and environmental metadata (image+traits+env). Metrics include Accuracy (Acc), Precision (Prec), Recall (Rec), F1 Score (F1), and Matthews Correlation Coefficient (MCC).}

\label{tab:multimodality_details}
\end{table*}

\end{document}